\documentclass[sigconf]{acmart}
\AtBeginDocument{%
  \providecommand\BibTeX{{%
    \normalfont B\kern-0.5em{\scshape i\kern-0.25em b}\kern-0.8em\TeX}}}

\copyrightyear{2022}
\acmYear{2022}
\setcopyright{acmcopyright}\acmConference[FAccT '22]{2022 ACM Conference on Fairness, Accountability, and Transparency}{June 21--24, 2022}{Seoul, Republic of Korea}
\acmBooktitle{2022 ACM Conference on Fairness, Accountability, and Transparency (FAccT '22), June 21--24, 2022, Seoul, Republic of Korea}
\acmPrice{15.00}
\acmDOI{10.1145/3531146.3533225}
\acmISBN{978-1-4503-9352-2/22/06}

\usepackage{algorithm}
\usepackage{algorithmic}
\usepackage{booktabs} % To thicken table lines
\usepackage{multirow,enumitem}
\usepackage{bbm}
\usepackage{subcaption}

\newcommand{\name}{{\textsc{FairPUL}}}
\newcommand{\hide}[1]{}
\newcommand*{\Scale}[2][4]{\scalebox{#1}{$#2$}}%

\DeclareMathOperator*{\argmin}{arg\,min}

\theoremstyle{definition}
\newtheorem{definition}{Definition}
\newtheorem{assumption}{Assumption}
\newtheorem{lemma}{Lemma}
\newtheorem{theorem}{Theorem}
\newenvironment{proofsketch}{%
  \renewcommand{\proofname}{Proof Sketch}\proof}{\endproof}

\begin{document}

\title{Fairness-aware Model-agnostic Positive and Unlabeled Learning}

\author{Ziwei Wu}
\email{ziweiwu2@illinois.edu}
\orcid{0000-0003-3999-4367}
\affiliation{
  \institution{University of Illinois at Urbana-Champaign}
  \city{Champaign}
  \country{USA}
}

\author{Jingrui He}
\email{jingrui@illinois.edu}
\orcid{0000-0002-6429-6272}
\affiliation{
  \institution{University of Illinois at Urbana-Champaign}
  \city{Champaign}
  \country{USA}
}

\renewcommand{\shortauthors}{Ziwei and Jingrui}


\begin{abstract}
With the increasing application of machine learning in high-stake decision-making problems, potential algorithmic bias towards people from certain social groups poses negative impacts on individuals and our society at large. In the real-world scenario, many such problems involve positive and unlabeled data such as medical diagnosis, criminal risk assessment and recommender systems. For instance, in medical diagnosis, only the diagnosed diseases will be recorded (positive) while others will not (unlabeled). 
Despite the large amount of existing work on fairness-aware machine learning in the (semi-)supervised and unsupervised settings, the fairness issue is largely under-explored in the aforementioned Positive and Unlabeled Learning (PUL) context, where it is usually more severe.
In this paper, to alleviate this tension, we propose a fairness-aware PUL method named \name. In particular, for binary classification 
over individuals from two populations, we aim to achieve similar true positive rates and false positive rates in both populations as our fairness metric. Based on the analysis of the optimal fair classifier for PUL, we design a model-agnostic post-processing framework, leveraging both the positive examples and unlabeled ones. Our framework is proven to be statistically consistent in terms of both the classification error and the fairness metric. Experiments on the synthetic and real-world data sets demonstrate that our framework outperforms state-of-the-art in both PUL and fair classification.
\end{abstract}


\maketitle

\section{Introduction}
\label{sec:intro}
Nowadays, machine learning systems are assisting, or in some cases even replacing, human decision making in an increasing number of application domains. Due to the profound impacts of these systems on individuals and our society at large, traditional performance metrics such as accuracy and precision are no longer the sole measure of success. In applications such as credit approval~\cite{twala2010multiple}, medical diagnosis~\cite{char2018implementing}, criminal risk assessment~\cite{perry2013predictive} and recommender systems~\cite{sweeney2013discrimination,wei2022comprehensive}, fairness must be carefully taken into account to ensure the absence of discrimination against certain social groups (e.g., women, blacks).
Recent years have witnessed a growing interest in fairness-aware machine learning to study the fairness issue. 
Various metrics of fairness for a predictive model have been studied in the literature~\cite{chouldechova2018frontiers}, including group fairness~\cite{chouldechova2017fair,zafar2017fairness,hardt2016}, individual fairness~\cite{jung2019eliciting,dwork2012fairness,dwork2020individual} and causal fairness~\cite{kusner2017counterfactual,yongkai2019pc}. Depending on the amount of label information available during training, researchers have developed a variety of algorithms to address the unfairness issue. For example, \cite{zafar2017fairness,hardt2016,zemel2013learning,song2019learning} focused on the traditional supervised learning setting where the learning algorithms have access to the class labels of all training examples; \cite{Chierichetti2017fairlet,kleindessner2019fair,chen2019proportionally} studied the unsupervised learning setting where no label information is available in the training set; and~\cite{chzhen2019,zhang2020fairness} mainly designed algorithms for the semi-supervised learning setting where only a small portion of the training examples contain class labels.

\hide{When learning predictive models from fully labeled data, machine learning algorithms have access to the class labels for all examples. Many algorithms have been developed to address fairness issues in this traditional supervised learning setting~\cite{zafar2017fairness,hardt2016,zemel2013learning,song2019learning}. Furthermore, considering absence of labels of samples, researchers explored the fairness issue in unsupervised learning~\cite{Chierichetti2017fairlet,kleindessner2019fair,chen2019proportionally} and semi-supervised learning settings~\cite{chzhen2019,zhang2020fairness}.}

In many real-world applications, the training data consists of only positive labeled examples and unlabeled ones, which cannot naturally fit in any of the aforementioned learning settings. For example, medical records usually only list which diseases a patient has been diagnosed with (i.e., positive samples) and they usually do not include which diseases a patient does not have. However, the absence of a diagnosis does not mean that the patient does not have the disease. Another example is scoring recidivism. When predicting criminal defendants' likelihood of re-offending, it is easy for us to collect part of the positive samples from the criminal records. For those who have not yet appeared in the criminal records, it is wrong to assume that they will not re-offend. They should therefore not be treated as negative examples but as unlabeled ones. In general, positive and unlabeled learning (PUL)~\cite{li2005learning} attempts to learn a classifier from this type of data. Existing (semi-)supervised and unsupervised methods can neither deal with positive-only labeled data nor make a full use of unlabeled data, and thus different PUL methods have been proposed~\cite{bekker2020learning}.

Despite the large amount of existing work on PUL~\cite{elkan2008learning,Plessis2014pul,Kiryo2017nonnegative,dhurandhar2020classifier}, the fairness issue in this setting has been largely under-explored. In applications such as medical diagnosis and recidivism scoring mentioned before, existing techniques can easily lead to inequality among different groups (e.g., black and white, female and male). The difference between fairness in PUL and in the other settings can be summarized in the following three folds: (1) Intuitively, the label imbalance will exacerbate the unfairness problem in PUL. (2) Empirically, methods proposed from supervised and semi-supervised settings cannot handle the unfairness issue in PUL as shown in Table \ref{tab:overall1}. (3) Theoretically, the 
properties (e.g. consistency) of other methods no longer hold in PUL due to the lack of negatively labeled data. Therefore, in this paper, we aim to bridge this gap and study the fairness-aware binary classification problem in PUL. To be more specific, we seek a classifier which minimizes the misclassification risk in PUL while satisfying the Equalized Odds / Equal Opportunity~\cite{hardt2016} fairness constraint. We derive the optimal fair classifier via recalibration of the Bayes regressor. This theoretical result motivates us to devise a generic post-processing framework named \name. Based on both positive and unlabeled examples, \name\ estimates the regression function and the unknown threshold to achieve the fairness criterion in the PUL setting. It enjoys the consistency property where it asymptotically satisfies the fairness criterion and its risk converges to the one of the theoretical optimal fair classifier. Extensive experiments on both the synthetic and the real-world data sets demonstrate \name's effectiveness.
Our main contributions can be summarized as follows:
\begin{itemize}
    \item To the best of our knowledge, this is the first work to systematically study the fairness issue in the positive and unlabeled learning setting.
    \item We derive the optimal fair classifier in PUL and propose a model-agnostic post-processing framework, which can accommodate different base models and enjoys the consistency property.
    \item Experiments on the synthetic and real-world data sets show that our framework performs favorably against state-of-the-art in both PUL and fair classification.
\end{itemize}

The rest of the paper is organized as follows. After a brief review of the related work in Section \ref{sec:related}, we present the problem definition in Section~\ref{sec:problemdef}. Section~\ref{sec:method} describes our proposed framework. The experimental results are discussed in Section~\ref{sec:exp}. Finally, we conclude the paper in Section~\ref{sec:con}.
\section{Related Work}
\label{sec:related}
% In this section, we briefly review the related work on fairness in classification as well as positive and unlabeled learning.
\subsection{Positive and Unlabeled Learning}
\label{sec:pulrelated}
Positive and Unlabeled Learning is a variant of the classical classification setup where the training data consists of only positive and unlabeled (PU) examples. 
It fits within the long-standing interest in developing learning algorithms that do not require fully supervised data. 
The PU data can originate from two scenarios. One is called the single-training-set scenario~\cite{elkan2008learning} where the data come from one single training set; the other is called the case-control scenario~\cite{ward2009presence} where the data come from two independently drawn datasets, one with all positive examples and one with all unlabeled examples. The single-training-set scenario has received substantially more attention in the literature~\cite{bekker2020learning}. Most PUL methods can be divided into the following three categories. Two-step techniques~\cite{he2018instance,ienco2016positive} first identify reliable negative examples and then conduct classical classification. Biased learning~\cite{shao2015laplacian,hsieh2015pu,claesen2015robust} considers PU data as fully labeled data with class label noise for the negative class. Class prior incorporation~\cite{elkan2008learning,Kiryo2017nonnegative,Plessis2014pul} modifies standard learning methods using the class prior and learns weights for all examples. However, none of the existing work in PUL explores the fairness issue. As shown in our experiments (Subsection \ref{sec:realexp}), current PUL methods exhibit unfairness against subgroups, which will result in potential discrimination in real-world applications.

\subsection{Fairness in Classification}
The existing studies on fairness in classification have focused on two key issues: how to formalize the fairness metric in classification, and how to design efficient algorithms that strike a desirable trade-off between classification performance and fairness.
To seek equality between different populations, a number of works have been proposed to control group fairness.
Equal Opportunity~\cite{hardt2016} requires the same true positive rates among groups and Equalized Odds~\cite{berk2018fairness} simultaneously considers false positive rates. Equalizing Disincentives~\cite{feldman2015certifying} requires the difference of two metrics to be equal across the groups. Using these metrics, a variety of fairness-aware algorithms in classification have been proposed under different learning settings. \cite{zafar2017fairness,hardt2016,zemel2013learning,agarwal2018reductions,song2019learning} focused on the supervised learning setting where the learning algorithms have access to the class labels of all training examples. Since they highly rely on supervised data, even if we adjust their decision boundary with a standard PUL technique, the unlabeled data still cannot contribute to algorithmic fairness explicitly. In a more realistic scenario where both labeled and unlabeled examples are given, \cite{bird2020fairlearn} devised a linear transformation pre-processing technique to remove the underlying discrimination. \cite{zhang2020fairness} incorporated the fairness constraint into the original training process. \cite{chzhen2019} post-processed the output conditional probability to improve fairness by recalibration with the unlabeled data. Although these methods do not rely on labeled data as much as the fully supervised methods, both positive and negative labeled samples are necessary for their algorithms. The most related work to us is \cite{chzhen2019} where they limit the fairness criterion to Equal Opportunity. However, the unlabeled examples can only be used to estimate the fairness threshold in their method. How to properly use the positive-only and unlabeled data in fair classification is still under-explored. In this paper, we instead explore how to make the best use of PU data to boost both classification performance and algorithmic fairness.

\section{Problem Definition}
\label{sec:problemdef}
We consider a fairness-aware binary classification task in the single-training-set scenario~\cite{elkan2008learning} as explained in Subsection~\ref{sec:pulrelated}.
The training examples consist of tuples $(X,S,L)$ where $X \in \mathbb{R}^d$ is a feature vector, $S \in \{0, 1\}$ is a binary sensitive attribute,
and $L \in \{0, 1\}$ represents whether the example is labeled ($L=1$) or not. Notice that we focus on one binary sensitive attribute in this paper, although the proposed technique can be naturally extended to multiple multi-class sensitive attributes (details in Subsection \ref{sub:optfair}).
The training examples are drawn randomly from distribution $p(X,S,L,Y)$, where $Y \in \{0,1\}$ is the ground-truth label. But for each tuple that is drawn, only $(X,S,L)$ is observed. 
% While $Y$ is always missing, information about it can be derived from the value of $L$ as only positive examples are labeled. In other words, if $L = 1$, then the example belongs to the positive class as $\mathbb{P}(Y = 1|L = 1) = 1$; if $L = 0$, the example can belong to either class. 
A classifier $g$ receives a pair $(X,S)$ as input, and outputs a binary prediction for the label. The set of all such functions from $\mathbb{R}^d \times \{0,1\}$ to $\{0,1\}$ is denoted by $\mathcal{G}$. For any classifier $g$, we denote its associated misclassification risk as $\mathcal{R}(g)$. An optimal fair classifier is then defined as:
$$ g^* = \arg\min_{g \in \mathcal{G}}\{ \mathcal{R}(g) : g \text{ is fair}\}$$

%\subsection{Fairness Definition}

Various definitions of fairness have been proposed so far~\cite{chouldechova2018frontiers}, but there is no consensus regarding which definition is universally the most appropriate. In this work, we employ the following group fairness metrics Equalized Odds and Equal Opportunity introduced in~\cite{hardt2016}:

\begin{definition}[Equalized Odds (EO)]
We say that a binary classifier $g(X,S)$ satisfies equalized odds with respect to $S$ and $Y$ if $\mathbb{P}\left(g(X, S) = 1 | Y = y, S = 1\right) = \mathbb{P}\left( g(X, S) = 1 | Y = y, S = 0\right), y \in \{0,1\} $.
\end{definition}

\begin{definition}[Equal Opportunity (EOP)]
We say that a binary classifier $g(X,S)$ satisfies equal opportunity with respect to $S$ and $Y$ if $\mathbb{P}\left(g(X, S) = 1 | Y = 1, S = 1\right) = \mathbb{P}\left( g(X, S) = 1 | Y = 1, S = 0\right)$.
\end{definition}

In the two fairness metrics, EO requires the same true positive rates and the same false positive rates across the sensitive groups. Compared to EO, EOP is a weaker notion which only requires the same true positive rates. The two fairness metrics have been used extensively in the literature either as a post-processing step~\cite{hardt2016} on a learned classifier or directly during training~\cite{donini2018empirical}. The motivation of these metrics and more discussion regarding the comparison with other fairness metrics can be found in~\cite{hardt2016,agarwal2018reductions,menon2018cost}. 

%\subsection{Risk Function}

% \subsection{Fair classification}
\section{The Proposed \name\ Framework}
\label{sec:method}
In this section, we introduce the model-agnostic post-processing framework \name\ for fairness-aware PUL.
\subsection{Labeling Mechanism in PUL}
As stated before, the unlabeled samples in PUL mainly come from the following two sources: (1) It is truly a negative example; (2) It is a positive example, but simply was not selected by the labeling mechanism. In order to enable learning with PU data, it is necessary~\cite{bekker2020learning} to make assumptions about either the labeling mechanism, the class distributions in the data, or both. We base our work on the most frequently used assumption~\cite{lee2003learning} in PUL:

\begin{assumption}[Selected Completely At Random (SCAR)]\label{assum:SCAR}
  Labeled examples are selected  completely at random, independent from their attributes, from the positive distribution. Formally speaking, $\mathbb{P}(L = 1|X, S, Y = 1) = \mathbb{P}(L = 1|Y = 1) \triangleq c$, where $c$ is called label frequency.
\end{assumption}

Under this assumption, the set of labeled examples are i.i.d. samples from the positive distribution. 
The label frequency $c$ plays an important role in the single-training-set scenario \cite{elkan2008learning} we consider here.
% The single-training-set scenario~\cite{elkan2008learning} we consider here allows the constant $c$ to be estimated in the following way. 
Define the regression function $f(X,S) := \mathbb{P}(L=1\mid X, S)$, i.e., the probability of an example $(X,S)$ being labeled. Then we have the following lemma regarding the constant $c$:

\begin{lemma}
\label{lemma:scar}
Suppose that the SCAR assumption holds. Then $\mathbb{P}(Y=1 \mid X, S) = f(X,S) / c$.
\end{lemma}

The proof can be found in Appendix \ref{prof:lemma1}. This lemma shows how we can obtain the probability of an example being positive from $f$. Next, we further incorporate fairness into the classification problem and show how to get the optimal fair classifier theoretically and empirically.

% For data distribution, the most simple and most naive assumption is to assume that the unlabeled examples all belong to the negative class. Despite the fact that this assumption obviously does not hold, it is often used in practice because it enables the use of standard machine learning methods for supervised binary classification~\cite{neelakantan2015compositional}. This naive assumption based method will serve as a baseline. Our work assumes the separability of positive and negative data distributions:
%\begin{assumption}[Separability]\label{assum:sep}
%  There exists a function in the considered hypothesis space that maps all the positive examples to a value that is higher or equal to a threshold $\tau$ and all negative examples to a value that is lower than $\tau$.
%\end{assumption}

\subsection{Optimal Fair Classifier in PUL}
\label{sub:optfair}
% For the sake of notation simplicity, we denote the two probabilities in EO as $P_1$ and $P_0$ respectively.
% $$P_1 := \mathbb{P}(g(X, S)=1 \mid Y=1, S=1)$$
% $$P_0 := \mathbb{P}(g(X, S)=1 \mid Y=1, S=0)$$
% In the PUL setting, based on Lemma~\ref{lemma:scar}, these two terms can be rewritten as follows:
% \begin{equation}
% \label{equ:p1p0}
% \begin{aligned}
% P_1=\frac{\mathbb{E}_{X \mid S=1}[g(X, 1) f(X, 1)]}{c\mathbb{P}(Y=1 \mid S=1)} \\
% P_0=\frac{\mathbb{E}_{X \mid S=0}[g(X, 0) f(X, 0)]}{c\mathbb{P}(Y=1 \mid S=0)}
% \end{aligned}
% \end{equation}
To obtain the optimal fair classifier, we study the following problem:
% \he{What is $\mathcal{G}$? You should have introduced this in Section 3 the equation regarding the optimal fair classifier.}
\begin{equation}
\label{equ:obj}
\min _{g \in \mathcal{G}}\{\mathcal{R}(g): \mathbb{P}\left(g(X, S) = 1 | Y = y, S = 1\right) = \mathbb{P}\left( g(X, S) = 1 | Y = y, S = 0\right), y \in \mathcal{Y}\}
\end{equation}

% \noindent Using weak duality we can write Equation (\ref{equ:obj}) as:
% \begin{equation}
% \label{equ:maxmin}
% \begin{aligned}
% &\min _{g \in \mathcal{G}} \max _{\lambda \in \mathbb{R}}\{\mathcal{R}(g)+\lambda(P_1 -P_0)\} \\
% & \geq \max _{\lambda \in \mathbb{R}} \min _{g \in \mathcal{G}}\{\mathcal{R}(g)+\lambda(P_1-P_0)\}
% % =:(* *)
% \end{aligned}
% \end{equation}

% \noindent Furthermore, regarding the misclassification risk, we have:
% \begin{align*}
% \mathcal{R}(g) =& \mathbb{P}(g \neq Y)\\
% =& \mathbb{P}(g=0, Y=1)+\mathbb{P}(g=1, Y=0) \\
% =& \mathbb{P}(g=1)+\mathbb{P}(Y=1)-2 \mathbb{P}(g=1, Y=1) \\
% =& \mathbb{P}(Y=1)- E_1 - E_0
% \end{align*}

% \noindent where $E_1$ and $E_0$ denote $\mathbb{E}_{X \mid S=1}[g(X, 1)(\frac{2}{c} f(X, 1)-1)] \mathbb{P}(S=1)$ and $\mathbb{E}_{X \mid S=0}[g(X, 0)(\frac{2}{c} f(X, 0)-1)] \mathbb{P}(S=0)$ respectively.
For EO, $\mathcal{Y} = \{0,1\}$. When $y=1$, the constraint requires the same true positive rates (TPR), that is $TPR^{(1)} = TPR^{(0)}$, where we use the superscript to identify the sensitive attribute $S$. Similarly, when $y=0$, the constraint requires the same false positive rates (FPR), that is $FPR^{(1)} = FPR^{(0)}$. For EOP, $\mathcal{Y} = \{1\}$ and it only requires $TPR^{(1)} = TPR^{(0)}$. In our later derivation, we will focus on EO while discussing how our results can easily accommodate EOP.

Using the misclassification risk $\mathcal{R}(g) = \mathbb{P}(g \neq Y)$ and based on Lemma \ref{lemma:scar},
we can solve problem (\ref{equ:obj}) and get the following result:

\begin{lemma}
The minimizer $g_{\boldsymbol{\lambda}}^*$ for every Lagrange multipliers $\boldsymbol{\lambda} = (\lambda_1, \lambda_2) \in \mathbb{R}^2$ is:
\begin{equation}
\label{equ:optimal}
\begin{aligned}
g_{\boldsymbol{\lambda}}^*(X,1) =& \mathbbm{1}_{\{\frac{f(X, 1)}{c}(1-\frac{\lambda_1}{\mathbb{P}(Y=1, S=1)}) + (1-\frac{f(X, 1)}{c}) (1-\frac{\lambda_2}{\mathbb{P}(Y=0, S=1)})\geq 0\}} \\
g_{\boldsymbol{\lambda}}^*(X,0) =& \mathbbm{1}_{\{\frac{f(X, 0)}{c}(1+\frac{\lambda_1}{\mathbb{P}(Y=1, S=0)}) - (1-\frac{f(X, 0)}{c}) (1+\frac{\lambda_2}{\mathbb{P}(Y=0, S=0)})\geq 0\}} 
\end{aligned}
\end{equation}
\end{lemma}

Here, $\lambda_1$ is the Lagrange multiplier for the same TPR constraint, and $\lambda_2$ is for the same FPR constraint. Since $g(X,S) \in \{0,1\}$, we can rewrite the Lagrange function of Problem (\ref{equ:obj}) into expressions of $g(X,S)$ and then get the above equations. The detailed derivation is shown in Appendix \ref{app:deri}. Note that by incorporating $c$ explicitly, our designed minimizer can boost performance and meanwhile reduce discrimination for PU data.

\noindent\textbf{Remark.} We consider one binary sensitive attribute in this problem for derivation simplicity. This can be naturally extended to accommodate multiple multi-class sensitive attributes (such as White, Black, Asian for race). We can break down the fairness into pairs of equations, add more equality constraints of EOD/EOP with respect to all sensitive attributes to the optimization problem in Equation (\ref{equ:obj}), and obtain a similar solution as Equation (\ref{equ:optimal}).

% On the other hand, in order to obtain the optimal value of $\lambda$ in the dual problem in Equation (\ref{equ:maxmin}), 
With Lemma \ref{equ:optimal}, the problem of finding the optimal fair classifier in PUL is equivalent to obtaining the optimal value of $\boldsymbol{\lambda}$ in Equation (\ref{equ:optimal}). To do this, we introduce an assumption on the regression function $\mathbb{P}(Y=1 \mid X, S)$, i.e., the probability of an example $(X,S)$ being positive. For notation simplicity, we abbreviate it as $p^+_{X,S}$.

\begin{assumption}\label{assum:conti}
The mapping $t \mapsto \mathbb{P}(p^+_{X,S} \leq t | S=s)$ is continuous on $(0,1)$.
% and $\mathbb{P}(p^+_{X,S} \geq / 1/2 | S=s) > 0$ for each $s \in \{0,1\}$.
\end{assumption}

% \textbf{Remark.} 
This Assumption requires that the random variable $p^+_{X,s}$ does not have atoms for each $s \in \{0,1\}$. This is proved achievable by many distributions \cite{yan2018binary,sadinle2019least,denis2020consistency}. 
% The second part requires that the random variable $p^+_{X,s}$ must surpass 1/2 with a non-zero probability for each $s \in \{0,1\}$. Consider the crime re-offending example in Section~\ref{sec:intro}. This requires that there exist criminal defendants from both black and white groups who are more likely not to re-offend.
Based on this assumption, we introduce the optimal $\lambda^*$:
\begin{lemma}
% \label{lemma:optimallambda}
The optimal $\boldsymbol{\lambda}^*$ for $g^*$ satisfies
% $\lambda^* \in [-2,2]$ and 
\begin{small}
\begin{equation}
\begin{aligned}
\label{equ:lambda}
\frac{\mathbb{E}_{X \mid S=1}[g_{\boldsymbol{\lambda}^*}^*(X,1)f(X, 1)]}{\mathbb{P}(Y=1 \mid S=1)} &= \frac{\mathbb{E}_{X \mid S=0}[g_{\boldsymbol{\lambda}^*}^*(X,0)f(X, 0)]}{\mathbb{P}(Y=1 \mid S=0)} \\
\frac{\mathbb{E}_{X \mid S=1}[(c-f(X, 1))(1-g_{\boldsymbol{\lambda}^*}^*(X,1))]}{\mathbb{P}(Y=0 \mid S=1)} &= \frac{\mathbb{E}_{X \mid S=0}[(c-f(X, 0))(1-g_{\boldsymbol{\lambda}^*}^*(X,0))]}{\mathbb{P}(Y=0 \mid S=0)} 
\end{aligned}
\end{equation}
\end{small}
\end{lemma}

\begin{proofsketch}
% The range of $\boldsymbol{\lambda}^*$ can be proved by contradiction based on the second part of Assumption~\ref{assum:conti}. 
Equation (\ref{equ:lambda}) is obtained with first-order optimality conditions, and its optimality can be proved based on Assumption~\ref{assum:conti}. The detailed proof is shown in Appendix \ref{prof:lemma2}.
\end{proofsketch}

\noindent\textbf{Remark.} The results in Lemma 2 are the minimizer under the fairness metric of EO, and Equations (\ref{equ:lambda}) show the constraints for the optimal $\boldsymbol{\lambda}^*$ under EO. To accommodate EOP, since we do not need to consider TNR, we can simply let $\lambda_2 = 0$, and the optimal $\boldsymbol{\lambda}^*$ only needs to satisfy the first equation in Equations (\ref{equ:lambda}). Notice that since $\lambda_2 = 0$ for EOP, the first equation is only the constraint for the optimal $\lambda_1$.

% Therefore, the pair $(\lambda^*, g_{\lambda^*}^*)$ is a solution to the dual problem in Equation~(\ref{equ:maxmin}) and we have $g^* = g_{\lambda^*}^*$.

% Finally, we derive the range of the optimal $\lambda^*$. Suppose that $2-\frac{\lambda^*}{\mathbb{P}(Y=1, S=1)} \leq 0$, then clearly $2+\frac{\lambda^*}{\mathbb{P}(Y=1, S=0)} > 0$. We can also have $g_{\lambda^*}^*(X,1) = 0$ according to Equation (\ref{equ:optimal}). Then the condition on $\lambda^*$ in Equation (\ref{equ:lambda}) can be derived as follows:
% \begin{align*}
%     0 &= \mathbb{E}_{X \mid S=0}[f(X, 0)\mathbf{1}_{\{1-\frac{f(X, 0)}{c}(2+\frac{\lambda^*}{\mathbb{P}(Y=1, S=0)})\leq 0\}}]\\
%     &\geq \frac{c \cdot \mathbb{P}(\frac{f(X, 0)}{c} \geq \frac{1}{2+\frac{\lambda^*}{\mathbb{P}(Y=1, S=0)})} | S=0)}{2+\frac{\lambda^*}{\mathbb{P}(Y=1, S=0)}}\\
%     &\geq \frac{c \cdot \mathbb{P}(p^+_{X,0} \geq \frac{1}{2} | S=0)}{2+\frac{\lambda^*}{\mathbb{P}(Y=1, S=0)}} >0
% \end{align*}
% where the second last inequality is based on Lemma~\ref{lemma:scar}, and the last inequality is due to Assumption~\ref{assum:conti}. This is clearly a contradiction. So the supposition does not hold, and thus we have $2-\frac{\lambda^*}{\mathbb{P}(Y=1, S=1)} > 0$.
% % \he{I don't see why this is the case.} 
% By similar way of reasoning, we also have $2+\frac{\lambda^*}{\mathbb{P}(Y=1, S=0)} > 0$. Combining these two inequalities, we have the following range of the optimal $\lambda^* \in [-2,2]$.

\subsection{\name\ for Fairness-aware PUL}
Based on the theoretical optimal classifier in Equations (\ref{equ:optimal}) and (\ref{equ:lambda}), we are now ready to introduce the proposed \name\ framework for empirically estimating the optimal fair classifier. Given an existing classifier, \name\ will modify its output for PUL data while reducing unfairness, and output the fair version of the classifier with a low misclassification risk.
% \he{\name\ is a post-processing framework, right? You should discuss its use in here. For example, say something like "It is a post-processing framework. Given an existing classifier, \name\ will take into consideration the fairness metrics, and outputs the fair-aware version of the classifier with a low misclassification risk."}
In the next subsections, any notation (e.g., $c$) with a hat denotes its estimated counterpart (e.g., $\hat{c}$).

In the PUL setting, the training data consists of two parts, the labeled (positive) data set $D_L$ and the unlabeled data set $D_U$. A classifier $\hat{f}$ can be constructed to estimate the probability of an example being labeled by treating $D_L$ as positive examples and $D_U$ as negative ones. 
 Let $V$ be the validation set that is drawn from the overall distribution in the same manner as the training set and $P$ be the subset of examples in $V$ that are labeled (and positive). We can estimate the label frequency $c$ as follows:
\begin{equation}
\label{equ:chat}
    \hat{c} = \frac{1}{n_P}\sum_{(x,s) \in P}\hat{f}(x,s)
\end{equation}
where $n_P$ is the cardinality of $P$. If $\hat{f}$ is trained well enough, $\hat{f}(x,s)$ provides a precise estimate of labeling probability. Since every example in $P$ is positive, we have $\hat{f}(x,s) \approx c$ based on Total Probability Theorem and SCAR assumption. Although this ideal case hardly holds in practice, the average estimation is a good choice for estimating the label frequency $c$ with low variance.

% Moreover, based on $\hat{f}$ and $\hat{c}$, we can construct the following estimator for the joint probability of the label and the sensitive attribute: 
% $$
% \hat{\mathbb{P}}(Y=1, S=s) = \hat{\mathbb{E}}_{X|S=s}[\hat{f}(X,s)] \hat{\mathbb{P}}(S=s) / \hat{c}
% $$

 Since the ideal $\boldsymbol{\lambda}^*$ should satisfy Equations (\ref{equ:lambda}), we minimize the difference between its two sides. Based on the rule of conditional probability, this is equivalent to minimizing unfairness defined as follows:

\begin{definition}[Unfairness]
For a binary classifier $g$, its unfairness $\Delta$ under different fairness metrics can be defined as:

Under EO: $$  \Delta_{EO}(g) = AOD(g) = \frac{1}{2}\big[|\text{TPR}^{(1)} - \text{TPR}^{(0)}| + |\text{FPR}^{(1)} - \text{FPR}^{(0)}| \big]$$

Under EOP:
$$  \Delta_{EOP}(g) = EOD(g) = |\text{TPR}^{(1)} - \text{TPR}^{(0)}|$$
\end{definition}

We can get their empirical versions by substituting all the unknown terms with their empirical estimators:

\begin{equation}
\label{equ:empiricaldelta}
\begin{aligned}
    &\Scale[0.92]{\hat{\Delta}_{EO}(g) =\frac{1}{2}\big[ |\frac{\hat{\mathbb{E}}_{X \mid S=1}[\hat{f}(X, 1)g(X,1)]}{\hat{\mathbb{E}}_{X|S=1}[\hat{f}(X,1)]} - \frac{\hat{\mathbb{E}}_{X \mid S=0}[\hat{f}(X, 0)g(X,0)]}{\hat{\mathbb{E}}_{X|S=0}[\hat{f}(X,0)]}|} \\
    &\Scale[0.92]{+ |\frac{\hat{\mathbb{E}}_{X \mid S=1}[(\hat{c}-\hat{f}(X, 1))(1-g(X,1))]}{\hat{c} -\hat{\mathbb{E}}_{X|S=1}[\hat{f}(X,1)]} - \frac{\hat{\mathbb{E}}_{X \mid S=0}[(\hat{c}-\hat{f}(X, 0)) (1-g(X,0))]}{\hat{c} - \hat{\mathbb{E}}_{X|S=0}[\hat{f}(X,0)]}|\big]}\\
    &\Scale[0.92]{\hat{\Delta}_{EOP}(g) =|\frac{\hat{\mathbb{E}}_{X \mid S=1}[\hat{f}(X, 1)g(X,1)]}{\hat{\mathbb{E}}_{X|S=1}[\hat{f}(X,1)]} - \frac{\hat{\mathbb{E}}_{X \mid S=0}[\hat{f}(X, 0)g(X,0)]}{\hat{\mathbb{E}}_{X|S=0}[\hat{f}(X,0)]}|}\\ %\Scale[0.92]
\end{aligned}
\end{equation}

Then we can obtain the estimator of $\boldsymbol{\lambda}^*$ as:

\begin{equation}
\hat{\boldsymbol{\lambda}} = \argmin_{\boldsymbol{\lambda}} \hat{\Delta}(\hat{g}^*_{\boldsymbol{\lambda}})
\end{equation}
where $\hat{\Delta}$ can be either $\hat{\Delta}_{EO}$ or $\hat{\Delta}_{EOP}$ to satisfy EO or EOP respectively. $\hat{g}^*_{\boldsymbol{\lambda}}$ is the estimated optimal fair classifier defined in a similar way as Equation (\ref{equ:optimal}) by replacing all the unknown terms with their empirical estimators.:
\begin{equation}
\label{equ:empiricalg}
\begin{aligned}
% \hat{g}_{\lambda}(X,1) &=\mathbbm{1}_{\{1-\frac{\hat{f}(X, 1)}{\hat{c}}(2-\frac{\lambda}{\hat{\mathbb{P}}(Y=1, S=1)})\leq 0\}} \\
% \hat{g}_{\lambda}(X,0) &=\mathbbm{1}_{\{1-\frac{\hat{f}(X, 0)}{\hat{c}}(2+\frac{\lambda}{\hat{\mathbb{P}}(Y=1, S=0)})\leq 0\}} \\
\hat{g}_{\boldsymbol{\lambda}}^*(X,1) =& \mathbbm{1}_{\{\frac{\hat{f}(X, 1)}{\hat{c}}(1-\frac{\lambda_1}{\mathbb{\hat{P}}(Y=1, S=1)}) + (1-\frac{\hat{f}(X, 1)}{\hat{c}}) (1-\frac{\lambda_2}{\mathbb{\hat{P}}(Y=0, S=1)})\geq 0\}} \\
\hat{g}_{\boldsymbol{\lambda}}^*(X,0) =& \mathbbm{1}_{\{\frac{\hat{f}(X, 0)}{\hat{c}}(1+\frac{\lambda_1}{\mathbb{\hat{P}}(Y=1, S=0)}) - (1-\frac{\hat{f}(X, 0)}{\hat{c}}) (1+\frac{\lambda_2}{\mathbb{\hat{P}}(Y=0, S=0)})\geq 0\}} 
\end{aligned}
\end{equation}
We adopt the Simulated Annealing strategy to search for the optimal $\hat{\boldsymbol{\lambda}}$.

% From the formulas it is easy to see that our procedure can be computed in polynomial time. 
Alg.~\ref{alg:overall} summarizes our proposed \name\ framework. It takes the training and validation sets as well as a base model $\hat{f}$ as input, and outputs a fairness-aware classifier with a low misclassification risk. The base regressor $\hat{f}$ is trained with $D_L$ and $D_U$. It can be any classifier that outputs the conditional probability of an example being labeled.
First, an estimator of $c$ can be learned on the validation set with $\hat{f}$ in Step 
1. Then we use Simulated Annealing to search for the best $\hat{\lambda}$ to minimize unfairness for PUL data in Step 2 and 3. We finally compute the empirical optimal fair classifier accordingly in Step 4.

\begin{algorithm}[t]
\caption{\name}
% \textbf {Input:} Support set $S_i$ of all source tasks $\mathcal{T}_i$; \\
\begin{algorithmic}[1]
\REQUIRE Labeled data set $D_L$, unlabeled dataset $D_U$;
% \he{are these data sets used in any of the following steps?} 
Validation set $V$; Base model $\hat{f}$;
\ENSURE $\hat{g}_{\hat{\lambda}}$
% \REQUIRE Base model $\hat{f}$ trained with $D_L$ and $D_U$;
% \FOR{each epoch}
% \STATE Update $\hat{f}$ with $D_L$ and $D_U$.\he{no need to mention multiple epochs here}
% \ENDFOR
\STATE Estimate $\hat{c}$ in $V$ with Equation (\ref{equ:chat}).
\STATE Compute terms in $\hat{\Delta}(\hat{g}_\lambda)$ defined in Equations (\ref{equ:empiricaldelta}) and (\ref{equ:empiricalg}) with $D_L$ and $D_U$.
\STATE Find $\hat{\lambda}$ to minimize $\hat{\Delta}(\hat{g}_\lambda)$ with Simulated Annealing.
\STATE Compute $\hat{g}_{\hat{\lambda}}$ in Equation (\ref{equ:empiricalg}).
\end{algorithmic}
\label{alg:overall}
\end{algorithm}

\subsection{Consistency of \name}
Consistency is a desired property for a classifier in asymptotic theory.
% \he{for estimating what?}. 
It guarantees that with the increase in the amount of data, the estimation will converge to the true value. In this subsection, we show that the proposed framework \name \ is consistent, that is, it asymptotically satisfies the fairness criterion and its risk converges to the one of the theoretical optimal fair classifier.

First of all, following~\cite{chzhen2019}, we make the following realistic assumptions on the estimator of $p^+_{X,S}$:

\begin{assumption}\label{assum:cons}
  The estimator $\hat{p}^+_{X,S}$ satisfies that, $\forall s \in \{0,1\}$,
  \begin{itemize}[leftmargin=*]
      \item $\mathbb{E}_{D}\mathbb{E}_{X|S=s}|p^+_{X,S} - \hat{p}^+_{X,S}| \to 0$ as $n_U, n_L \to \infty$, where $n_U$ and $n_L$ denote the number of examples in $D_U$ and $D_L$ respectively.
      \item There exists a sequence $c_{U,L} > 0$ satisfying $\frac{1}{c_{U,L}\sqrt{N}} = o_{U,L}(1)$ and $c_{U,L}= o_{U,L}(1)$ such that $\mathbb{E}_{X|S=s}[\hat{p}^+_{X,S}] \geq c_{U,L}$ almost surely.
      \item The mapping $t \mapsto \mathbb{P}(\hat{p}^+_{X,s} \leq t | S=s)$ is continuous on $(0,1)$ almost surely.
  \end{itemize}
\end{assumption}

% \noindent\textbf{Remark.} 
The first part of the assumption requires the estimator to be consistent in $l_1$ norm. This can be achieved by a variety of estimations for different regression functions as shown in the literature~\cite{devroye1978uniform,audibert2007fast,van2008high}. The second part means that $\mathbb{E}_{X|S=s}[\hat{p}^+_{X,S}]$ is lower bounded by a certain positive term which vanishes as $n_U$ and $n_L$ go to infinity. This can be easily achieved by slightly modifying any existing consistent estimator. 
% For instance, for a given estimator $\Tilde{p}^+_{X,S}$, one can let $\hat{p}^+_{X,S} = max\{\Tilde{p}^+_{X,S}, c_{U,L}\}$. 
% On the one hand, it naturally satisfies the second assumption. On the other hand, for a consistent $\Tilde{p}^+_{X,S}$, using triangle inequality we can have $\mathbb{E}_{D}\mathbb{E}_{X|S=s}|p^+_{X,S} - \hat{p}^+_{X,S}| \leq \mathbb{E}_{D}\mathbb{E}_{X|S=s}|p^+_{X,S} - \Tilde{p}^+_{X,S}| + \mathbb{E}_{D}\mathbb{E}_{X|S=s}|\Tilde{p}^+_{X,S} - \hat{p}^+_{X,S}| \leq \mathbb{E}_{D}\mathbb{E}_{X|S=s}|p^+_{X,S} - \Tilde{p}^+_{X,S}| + c_{U,L} \to 0$, which means that $\hat{p}^+_{X,S}$ is still consistent.
The last part is similar to Assumption \ref{assum:conti}. 
% This can be achieved for any estimator by adding small noise which  decreases with $n_U$ and $n_L$. It allows us to show that the empirical unfairness of the proposed framework is small or even zero in some cases, which is crucial for proving the consistency of \name.

Based on these realistic assumptions on $p^+_{X,S}$, next we establish the statistical consistency of \name.

\begin{theorem}[Asymptotic properties] \name~ satisfies:

% $\lim_{n_U,n_L \to \infty} \mathbb{E}\left[\Delta(\hat{g})\right] \to 0, \lim_{n_U,n_L \to \infty} \mathbb{E}\left[\mathcal{R}(\hat{g})\right] \leq \mathcal{R}(g*) $

$\lim_{n_U,n_L \to \infty} \mathbb{E}\left[\Delta(\hat{g})\right] = 0, \lim_{n_U,n_L \to \infty} \mathbb{E}\left[\mathcal{R}(\hat{g})\right] = \mathcal{R}(g*) $

\end{theorem}

\begin{proofsketch}
To prove asymptotic optimality (the second part), we introduce an intermediate estimator and use it to upper bound the excess risk with triangle inequality. The upper bound converges to zero based on the first two parts in Assumption \ref{assum:cons}. For asymptotic fairness (the first part), we upper bound the unfairness with its empirical version which converges to zero. Detailed proof can be found in Appendix \ref{prof:theo1}. Recall that $\hat{p}^+_{X,S} = \hat{f}(X,S) / \hat{c}$. Its consistency as $n_U, n_L \to \infty$ is vital to the consistency of our framework in PUL.
\end{proofsketch}

\section{Experiments}
\label{sec:exp}
To evaluate the effectiveness of \name, we conduct extensive experiments to answer the following research questions:
\begin{itemize}[leftmargin=*]
    \item \textbf{RQ1}: How does \name\ perform compared with state-of-the-art in fair classification and PUL?
    \item \textbf{RQ2}: How does \name\ for post-processing compare with in-processing and pre-processing methods?
    \item \textbf{RQ3}: How do unlabeled examples affect \name?
\end{itemize}

\subsection{Experiments on Synthetic Data.}

\subsubsection{Experimental setup}
The aim of the synthetic experiment is to study the behavior of \name\ in comparison with other methods with the base model of linear logistic regression (Lin.LR), in terms of both classification performance and fairness.
To this end, we generate a synthetic binary classification data set with two sensitive groups following Donini et al.~\cite{donini2018empirical}. For each group in the class 0 and for the group $a$ in the class 1, we generate 1,000 examples for training and the same number for testing. For the group $b$ in the class 1, we generate 200 examples for training and the same number for testing. Each set of examples is sampled from a 2-dimensional isotropic Gaussian distribution with different mean $\mu$ and variance $\sigma^2$ : (i) Group $a$, Label 1: $\mu$ = (-1, -1), $\sigma^2$ = 0.8; (ii) Group $a$, Label 0: $\mu$ = (1, 1), $\sigma^2$ = 0.8; (iii) Group $b$, Label 1: $\mu$ = (-0.5, -0.5), $\sigma^2$ = 0.5; (iv) Group $b$, Label 0: $\mu$ = (0.5, 0.5), $\sigma^2$ = 0.5. When a standard machine learning method is applied to this synthetic data set, the generated model is unfair with respect to the group $b$, in that the classifier tends to negatively classify the examples in this group. We search in $[0.01, 0.1, 1, 10, 100]$ for the best regularization parameter $C$. We generate the validation set from the training set via holdout validation and the holdout ratio is set to 0.2.

\subsubsection{Baselines}
\label{sec:baselines}
%To compare \model~with others, the following methods are adopted as baselines:
To study and compare the performance of \name, we first adopt two classic methods as baselines:
\begin{itemize}[leftmargin=*]
\item \textbf{Oracle:} A method using the fully labeled training set (all examples are labeled either positive or negative). Its results correspond to the fully supervised setting.
\item \textbf{Na\"ive:} A method using the labeled examples as positive ones and treating unlabeled examples as negative.
\end{itemize}

Since this is the first work on the fairness issue in the PUL setting, we further compare \name\ with the following two types of existing work.
For fairness work:
\begin{itemize}[leftmargin=*]
\item \textbf{Agarwal~\cite{agarwal2018reductions}:} An in-processing method reducing fair classification to cost-sensitive classification problems and yielding a randomized classifier with the lowest error subject to the desired constraints.
\item \textbf{Hardt~\cite{hardt2016}:} A post-processing method which takes as input an existing classifier and the sensitive feature, and derives a monotone transformation of the prediction to enforce the specified fairness constraints.
\item \textbf{Chzhen~\cite{chzhen2019}:} A post-processing method which recalibrates the Bayes classifier by a group-dependent threshold to minimize unfairness.
\end{itemize}

All the above fairness baselines adopt the similar fairness metric as we do.
For PUL work:
\begin{itemize}[leftmargin=*]
\item \textbf{uPU/wPU~\cite{elkan2008learning}:} Unbiased PUL methods under the SCAR assumption by reweighting the examples. Both unweighted (uPU) and weighted (wPU) versions were proposed.
\item \textbf{nnPU~\cite{Kiryo2017nonnegative}:} A designed non-negative risk estimator for PUL which can be trained on flexible neural networks when minimized.
\item \textbf{BaggingSVM~\cite{mordelet2014bagging}:} A  bootstrap-bagging-based model which iteratively trains multiple classifiers to discriminate the known positive examples from random subsamples of the unlabeled set, and averages their predictions. 
% \item \textbf{Biased-SVM:} 
\end{itemize}

\subsubsection{Results (RQ1)}

\begin{figure}[t]
\centering
\begin{subfigure}{7cm}
  \centering
  \includegraphics[width=7cm]{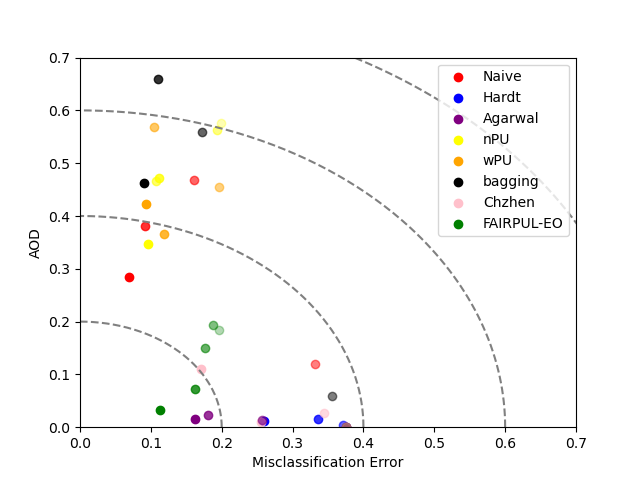}
  \caption{\name-EO}
  \label{fig:sub1}
\end{subfigure} \\
\begin{subfigure}{7cm}
  \centering
  \includegraphics[width=7cm]{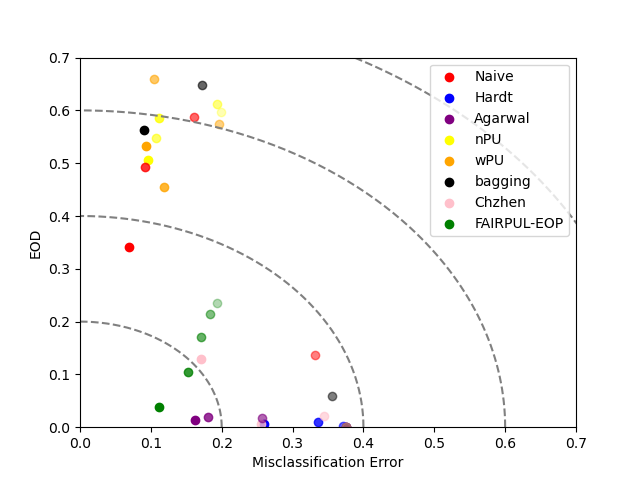}
  \caption{\name-EOP}
  \label{fig:sub2}
\end{subfigure}
\caption{Test classification error, AOD and EOD under different labeling rates. The dots with higher transparency correspond to the results under lower labeling rates.}
\Description{Test classification error, AOD and EOD under different labeling rates.}
\label{fig:synthe}
\end{figure}

We compare the test classification error and the fairness metrics under different labeling rates in $[1.0, 0.8,0.6,0.4,0.2]$ of the baselines and our proposed \name. For \name\ with EO constraint, the fairness metric is AOD. For \name\ with EOP constraint, the fairness metric is EOD. The base model for all the methods is Lin.LR.  We report the average results of misclassification error, AOD and EOD of 5 independent trials in Fig.~\ref{fig:synthe}. The closer the dot is to the origin, the more fair and accurate the model is. From the figures we can see that for all the methods, their classification performances drop as the labeling rate decreases. Under high labeling rates ($\geq 0.8$), our framework achieves much lower AOD and EOD (i.e., higher level of fairness) while maintaining a good level of accuracy; under low labeling rates, our framework achieves much better performance in both classification accuracy and fairness. Note that even though some methods achieve zero AOD/EOD when the labeling rate is 0.2, it is not an ideal model we are seeking. In this case, the model simply predicts every example to be negative, showing "fake" fairness. What we are seeking is maintaining task-specific performance and reducing discrimination simultaneously.

\subsection{Experiments on Real Data.}
\label{sec:realexp}

\subsubsection{Data sets and experimental setup}

To further study how \name \ performs, we conduct experiments on three publicly available real-world data sets. In all the experiments, to obtain reliable estimates of classification performance and fairness, we repeatedly randomly split each data set into training (70\%) and test (30\%) sets 10 times, and report the averages and standard deviations of the metrics over different independent runs. 
% Note that in each run, the selected positive examples and the data set splits are kept identical for all the competitors. 
Following \cite{gong2019large}, to realize the PUL setting, for each training set, we randomly select 90\% positive examples as labeled and leave the remaining 10\% positive examples as well as all negative examples as unlabeled. We split up the training set for holdout validation in \name\ and the holdout ratio is set to 0.2. 
% \textcolor{red}{[fair comparison]}
% \he{If only your framework uses the validation set, and no other methods use this set, the reviewers may say that this is not a fair comparison. Maybe you could let the other methods use the validation set in some way?}

\begin{itemize} %[leftmargin=*]
\item COMPAS recidivism data \cite{compas}.
% \footnote{Available at https://github.com/propublica/compas-analysis}. 
It includes 5278 records with 47\% positive examples. The task is to predict recidivism from someone's criminal history, jail and prison time, demographics, and COMPAS risk scores, with race as the protected sensitive attribute restricted to black (about 40\%) and white defendants. 
\item German Credit \cite{german}.
% \footnote{ Available at https://archive.ics.uci.edu/ml/machine-learning-databases/statlog/german/}. 
It includes 1000 examples where 30\% are positive. The task is to classify people as good or bad credit risks by features related to the economical situation, with gender as the sensitive attribute restricted to female (about 31\%) and male.
\item Drug \cite{fehrman2017five}. It comprises 1885 records of human subjects, and for each subject, it provides five demographic features, seven features measuring personality traits and 18 features each of which describes the subject’s last use of a certain drug. We choose the use of heroin here, where 80\% of the subjects have never used heroin. We restrict the sensitive attribute race to black (about 9\%) and white.
\end{itemize}

{
\begin{table*}[t]
\caption{Average results and standard deviations on real data of 10 runs. The labeling rate is 90\%. The top 2 results under each metric are marked bold.}
\Description{Average results and standard deviations on real data of 10 runs.}
\centering
% \fontsize{6.5pt}{6.5pt}\selectfont
% \vspace{-0.2cm}
\fontsize{6.8pt}{6.8pt}\selectfont
\begin{tabular}{l|c|c|c|c|c|c|c|c|c} 
\toprule
\multirow{2}{*}{Models} & \multicolumn{3}{c|}{German} &\multicolumn{3}{c|}{COMPAS} &\multicolumn{3}{c}{Drug}\\
\cmidrule{2-10}
 &  F1 & AOD &EOD & F1 & AOD&EOD & F1 & AOD&EOD\\
\midrule
 Oracle &0.565$\pm$0.051 &0.053$\pm$0.033&0.061$\pm$0.042& 0.604$\pm$0.018& 0.194$\pm$0.020&0.219$\pm$0.032 &0.777$\pm$0.021&0.130$\pm$0.024&0.041$\pm$0.036\\
 Na\"ive &0.471$\pm$0.048 & 0.068$\pm$0.024&0.099$\pm$0.052 &0.529$\pm$0.058 & 0.180$\pm$0.031&0.237$\pm$0.045&0.753$\pm$0.019&0.142$\pm$0.031&0.086$\pm$0.049\\
Agarwal & 0.470$\pm$0.033& 0.053$\pm$0.026&0.077$\pm$0.051& 0.467$\pm$0.042 &0.032$\pm$0.021& 0.040$\pm$0.036& 0.743$\pm$0.031&0.123$\pm$0.019&0.065$\pm$0.033\\
Hardt & 0.449$\pm$0.058&0.059$\pm$0.025&0.093$\pm$0.046& 0.409$\pm$0.062&0.034$\pm$0.023& 0.034$\pm$0.029&0.756$\pm$0.022&0.111$\pm$0.033&0.075$\pm$0.042\\ 
Chzhen &0.393$\pm$0.069&0.048$\pm$0.018&\textbf{0.053$\pm$0.028}& 0.486$\pm$0.047 &0.036$\pm$0.012& 0.032$\pm$0.020&0.755$\pm$0.020&0.108$\pm$0.024&0.041$\pm$0.027\\
uPU &0.574$\pm$0.022  &0.097$\pm$0.045&0.054$\pm$0.027& 0.644$\pm$0.012 & 0.031$\pm$0.024 &0.016$\pm$0.009&\textbf{0.871$\pm$0.026}&0.143$\pm$0.060&0.066$\pm$0.043\\
wPU & 0.534$\pm$0.131 &0.050$\pm$0.022&0.066$\pm$0.039& 0.612$\pm$0.071 & 0.188$\pm$0.035 &0.245$\pm$0.048 &0.766$\pm$0.027&0.147$\pm$0.016&0.074$\pm$0.049\\
 Bagging & 0.544$\pm$0.056 &0.054$\pm$0.034&0.062$\pm$0.052& \textbf{0.649$\pm$0.013} &0.215$\pm$0.031&0.252$\pm$0.038&0.813$\pm$0.020&0.196$\pm$0.041&0.074$\pm$0.051\\
 \name-EO & \textbf{0.578$\pm$0.023}&\textbf{0.045$\pm$0.018}&0.056$\pm$0.031& 0.646$\pm$0.012 & \textbf{0.005$\pm$0.003} & \textbf{0.004$\pm$0.002}&\textbf{0.871$\pm$0.021} &\textbf{0.056$\pm$0.023}&\textbf{0.024$\pm$0.013}\\
 \name-EOP &\textbf{0.580$\pm$0.020} &\textbf{0.042$\pm$0.029}&\textbf{0.034$\pm$0.027}& \textbf{0.648$\pm$0.013}& \textbf{0.006$\pm$0.004}&\textbf{0.003$\pm$0.002}&\textbf{0.897$\pm$0.023}&\textbf{0.061$\pm$0.048}&\textbf{0.022$\pm$0.010}\\
\bottomrule
% \toprule
% \multirow{2}{*}{Models} & \multicolumn{2}{c|}{German} &\multicolumn{2}{c|}{COMPAS} &\multicolumn{2}{c}{Drug}\\
% \cmidrule{2-7}
%  &  F1 & EOD & F1 & EOD & F1 & EOD\\
% \midrule
%  Oracle & 0.559$\pm$0.049& 0.078$\pm$0.062& 0.604$\pm$0.016& 0.219$\pm$0.032 &&\\
%  Na\"ive &0.459$\pm$0.043&0.087$\pm$0.056 &0.528$\pm$0.059 &0.237$\pm$0.045&&\\
% Agarwal & 0.418$\pm$0.059& 0.073$\pm$0.049& 0.467$\pm$0.042& 0.033$\pm$0.032&& \\
% % acc: 0.743,0.022; acc:0.624,0.018
% Hardt &0.436$\pm$0.057 & 0.114$\pm$0.039&0.409$\pm$0.062& 0.036$\pm$0.028&&\\ %ACC:german: 0.745,0.025 compas 0.596,0.015
% Chzhen &0.393$\pm$0.069 & \textbf{0.041$\pm$0.034} &0.486$\pm$0.047& \textbf{0.032$\pm$0.020}&&\\
% uPU &  \textbf{0.566$\pm$0.022} &0.076$\pm$0.045& 0.644$\pm$0.012& 0.116$\pm$0.019&&\\
% wPU &  0.277$\pm$0.131 & 0.066$\pm$0.039& 0.534$\pm$0.071 & 0.245$\pm$ 0.048 &&\\
%  Bagging &  0.545$\pm$0.040 & 0.079$\pm$0.060 & \textbf{0.650$\pm$0.012} & 0.252$\pm$0.034&&\\
%  \name-EOD & \textbf{0.565$\pm$0.020} &\textbf{0.034$\pm$0.033} & \textbf{0.646$\pm$0.012} & \textbf{0.006$\pm$0.005}&&\\
% \bottomrule
\end{tabular}
\label{tab:overall1}
\end{table*}
}

We compare \name\ with the baselines described in Subsection \ref{sec:baselines} with the base model of Linear Support Vector Machine (Lin.SVM). Since previous fairness works cannot be naturally adapted to the PUL setting, the fairness baselines we use here all follow the na\"ive method to transform PUL to the traditional learning setting. 
% The hidden layers sizes in MLP are set to $(8, 16, 2)$ for COMPAS and $(24, 48, 2)$ for German Credit. 
The hyper-parameters of every algorithm have been carefully tuned to achieve the best classification performance, and the details can be found in Appendix \ref{app:hyper}.
% with different base models, including Linear Support Vector Machine (Lin.SVM), Lin.LR, Support Vector Machine with polynomial kernel (SVM) and Multilayer Perceptron (MLP).

\subsubsection{Metrics}
We compare our framework with the baselines using the metrics of F1 score for evaluating the model performance, and AOD and EOD for evaluating the fairness level. It is worth noting that most previous works on fairness simply use classification accuracy to evaluate the models' performance. However, higher accuracy does not necessarily mean a better classification ability, especially on imbalanced data sets like German Credit and drug. Therefore, we use the F1 score instead.

\subsubsection{Results (RQ1)}

 The comparison of baselines and \name\ is summarized in Table~\ref{tab:overall1}.
From the results we have the following observations:

\begin{itemize}[leftmargin=*]
    \item The na\"ive and PUL baselines show high AOD and EOD, and most often higher than the oracle method. It demonstrates the unfairness problem in the PUL setting, which is usually more severe than in the supervised learning setting as we discussed before.
    % \item The na\"ive method and the fairness-aware methods based on it show unsatisfying classification performance. This indicates the necessity of properly treating and leveraging the unlabeled examples in the PUL setting.
    \item Fair classification methods often obtain lower AOD and EOD, and PUL methods often obtain higher F1 scores. However, none of them can perform well on both metrics. That is, existing fairness works cannot ensure model performance in the PUL setting, while existing PUL methods cannot ensure fairness.
    \item \name \ methods achieve both high F1 scores and low AOD and EOD. This demonstrates that our framework strikes a good trade-off between classification performance and fairness. \name \ even beats the oracle model. This may be due to the imbalance of the data sets, where \name\ resolves it via proper estimation of the label frequency.
    \item \name-EOP often achieves higher F1 scores than \name-EO under comparable AOD and EOD. This is because EOP is a weaker fairness constraint compared with EO, and thus it typically allows for stronger task performance.
\end{itemize}

% 100% alga: f1: 0.541(0.050)& DEO: 0.051(0.044) & ACC: 0.761(0.022)

We report the results of the methods with different base models on COMPAS in Table~\ref{tab:overall} in Appendix \ref{app:exp}. Similar observations can be drawn, which demonstrate that our proposed \name\ can always strike a good trade-off between classification performance and fairness, and generalize well to different base models.
{
\begin{table}[t]
\caption{Post-processing vs In-processing/Pre-processing. The best 2 results for each metric are marked bold.} 
\Description{Post-processing vs In-processing/Pre-processing.}
\centering
\fontsize{7pt}{7pt}\selectfont
\begin{tabular}{c|c|c|c|c|c} 
\toprule
Rates& \multicolumn{2}{c|}{Models}  &  F1 & AOD & EOD\\
\midrule
\multirow{8}{*}{100\%}& \multirow{4}{*}{Lin.SVM} & Na\"ive &  0.604$\pm$0.018& 0.194$\pm$0.020&0.219$\pm$0.032\\
~ & &In & 0.588$\pm$0.014 &0.030$\pm$0.026 &0.039$\pm$0.021 \\
~ & &Pre &  0.598$\pm$0.015 & 0.076$\pm$0.025 &0.098$\pm$0.027\\
~& &\name-EO & \textbf{0.649$\pm$0.013} & \textbf{0.004$\pm$0.003}& \textbf{0.005$\pm$0.004}\\ 
~& &\name-EOP & \textbf{0.650$\pm$0.014} & \textbf{0.008$\pm$0.004}&\textbf{0.004$\pm$0.002}\\ 
\cmidrule{2-6}
~& \multirow{4}{*}{Lin.LR} & Na\"ive & 0.621$\pm$0.015 & 0.236$\pm$0.033&0.293$\pm$0.045\\
~& &In & 0.595$\pm$0.011& 0.042$\pm$0.022&0.038$\pm$0.026\\ 
~& &Pre & 0.590$\pm$0.016& 0.054$\pm$0.029& 0.047$\pm$0.027\\ 
~& &\name-EO & \textbf{0.657$\pm$0.024}&\textbf{0.014$\pm$0.010}& \textbf{0.013$\pm$0.006}\\ 
~& &\name-EOP &\textbf{0.660$\pm$0.013} & \textbf{0.018$\pm$0.005}&\textbf{0.012$\pm$0.009}\\ 
\midrule
\multirow{8}{*}{90\%}& \multirow{4}{*}{Lin.SVM} & Na\"ive & 0.529$\pm$0.058 & 0.180$\pm$0.031&0.237$\pm$0.045\\
~ & &In & 0.467$\pm$0.042 &0.032$\pm$0.021& 0.040$\pm$0.036\\
~ & &Pre &  0.429$\pm$0.054 &0.028$\pm$0.016 &0.038$\pm$0.029\\
~& &\name-EO & \textbf{0.646$\pm$0.012} & \textbf{0.005$\pm$0.003} & \textbf{0.004$\pm$0.002}\\ 
~& &\name-EOP & \textbf{0.648$\pm$0.013}& \textbf{0.006$\pm$0.004}&\textbf{0.003$\pm$0.002}\\
\cmidrule{2-6}
~& \multirow{4}{*}{Lin.LR} & Na\"ive & 0.552$\pm$0.023 & 0.240$\pm$0.048&0.311$\pm$0.054\\
~& &In &  0.507$\pm$0.019& 0.032$\pm$0.021 &0.037$\pm$0.036\\
~& &Pre &  0.495$\pm$0.019& 0.049$\pm$0.030& 0.053$\pm$0.024\\ 
~& &\name-EO & \textbf{0.656$\pm$0.016}&\textbf{0.015$\pm$0.008}&\textbf{0.013$\pm$0.010}\\ 
~& &\name-EOP &\textbf{0.658$\pm$0.012} &\textbf{0.016$\pm$0.007} &\textbf{0.009$\pm$0.008}\\
\bottomrule
\end{tabular}
\label{tab:invspost}
\end{table}
}

{
\begin{table}[ht]
\caption{Comparison results of different labeling rates.} 
\Description{Comparison results of different labeling rates.} 
\vspace{-0.2cm}
\centering
\fontsize{7.5pt}{7.5pt}\selectfont
% \vspace{-0.2cm}
\begin{tabular}{c|c|c|c|c|c} 
\toprule
{Rates}& \multicolumn{2}{c|}{Models}  & F1 & AOD&EOD\\
\midrule
\multirow{4}{*}{100\%}& \multirow{2}{*}{Lin.SVM}& Na\"ive & 0.604$\pm$0.018& 0.194$\pm$0.020&0.219$\pm$0.032\\
~& &\name & 0.650$\pm$0.014 & 0.008$\pm$0.004&0.004$\pm$0.002\\ 
\cmidrule{2-6}
~& \multirow{2}{*}{Lin.LR}& Na\"ive & 0.621$\pm$0.015 & 0.236$\pm$0.033&0.293$\pm$0.045\\
~& &\name &0.660$\pm$0.013 & 0.018$\pm$0.005&0.012$\pm$0.009\\ 
\midrule
\multirow{4}{*}{90\%}& \multirow{2}{*}{Lin.SVM}& Na\"ive & 0.529$\pm$0.058 & 0.180$\pm$0.031&0.237$\pm$0.045\\
~& &\name &0.648$\pm$0.013& 0.006$\pm$0.004&0.003$\pm$0.002\\
\cmidrule{2-6}
~& \multirow{2}{*}{Lin.LR}& Na\"ive & 0.552$\pm$0.023 & 0.240$\pm$0.048&0.311$\pm$0.054\\
~& &\name &0.658$\pm$0.012 &0.016$\pm$0.007 &0.009$\pm$0.008\\
\midrule
\multirow{4}{*}{80\%}& \multirow{2}{*}{Lin.SVM}& Na\"ive &0.274$\pm$0.024&0.090$\pm$0.011&0.126$\pm$0.023\\
~& &\name &0.646$\pm$0.012&0.006$\pm$0.005&0.004$\pm$0.003\\
\cmidrule{2-6}
~& \multirow{2}{*}{Lin.LR}& Na\"ive &0.401$\pm$0.018& 0.142$\pm$0.021 &0.205$\pm$0.031\\
~& &\name &0.655$\pm$0.011&0.015$\pm$0.007&0.007$\pm$0.007\\
\midrule
\multirow{4}{*}{50\%}& \multirow{2}{*}{Lin.SVM}& Na\"ive & 0.000$\pm$0.000 &0.000$\pm$0.000&0.000$\pm$0.000\\
~& &\name & 0.644$\pm$0.012 &0.009$\pm$0.007& 0.006$\pm$0.005\\
\cmidrule{2-6}
~& \multirow{2}{*}{Lin.LR}& Na\"ive &0.068$\pm$0.015&0.022$\pm$0.009&0.037$\pm$0.016\\
~& &\name &0.652$\pm$0.014&0.017$\pm$0.007&0.008$\pm$0.004\\
\bottomrule
\end{tabular}
\vspace{-0.3cm}
\label{tab:rates}
\end{table}
}
\subsection{Post-processing vs. In-/Pre-processing (RQ2)}
In this section, we compare the post-processing \name\ with the in-processing method proposed by \cite{agarwal2018reductions} and the pre-processing method in~\cite{bird2020fairlearn}. The in-processing method is introduced in Section~\ref{sec:realexp}. The pre-processing method is a widely-used technique which transforms the non-sensitive features to remove their correlation with the sensitive feature while retaining as much information as possible. We report the average results of 10 independent runs on COMPAS data set under labeling rates of 1.0 and 0.9 in Table~\ref{tab:invspost}. Results on German can be found in Appendix \ref{app:exp}.
In the fully labeled setting, the in-processing and pre-processing methods sacrifice much classification performance for better fairness. \name\ achieves a good balance instead.
Under labeling rate 90\%, \name\ significantly outperforms all the baselines in both metrics. 
% Moreover, the in-processing and pre-processing methods are highly affected by labeling rates, indicating that they cannot accumwodate PUL data.
Compared with the in-processing method, \name\ only needs black-box access to the predictions and sensitive attribute information without requiring access to the actual algorithms and ML models. While the pre-processing method only needs to transform the data set before the actual model takes effect, it leads to classifiers that still exhibit substantial unfairness in practice. So our proposed \name\ is more flexible and also effective in fairness-aware PUL.

\subsection{In-depth Study}
\subsubsection{Effect of labeling rates (RQ3)}

% Following Gong et al.~\cite{gong2019large}, we transform the fully supervised data sets into the PUL setting. 
We test labeling rates of [100\%, 90\%, 80\%, 50\%] on the base models Lin.SVM and Lin.LR. Changing the labeling rates affects the number of positive and unlabeled examples simultaneously. As we have observed before, \name-EOP often achieves a better trade-off between fairness and classification compared with \name-EO, so we only report the results of \name-EOP here.   Average results of 10 runs on COMPAS data set are shown in Table~\ref{tab:rates}. Additional results on German are in Appendix \ref{app:exp}. From the table we can see that \name\ always achieves higher F1 scores and lower AOD and EOD under different labeling rates compared with the na\"ive method. 
% It demonstrates \name's ability to improve classification and decrease unfairness by properly treating and leveraging unlabeled data.
Although \name's classification performance drops with the decrease in labeling rates, it shows much more stable performance and manages to always maintain a good level of fairness. Even under the labeling rate as low as 50\%, where the na\"ive method fails and simply predicts every example as negative, \name\ can achieve even comparable performance with na\"ive method in the fully labeled setting. This means that \name\ can achieve satisfying performance with much fewer labeled examples. It is a very desirable property in practice where labeled data is often time-consuming and expensive to obtain.

\begin{figure}[t]
\centering
\includegraphics[width=6cm]{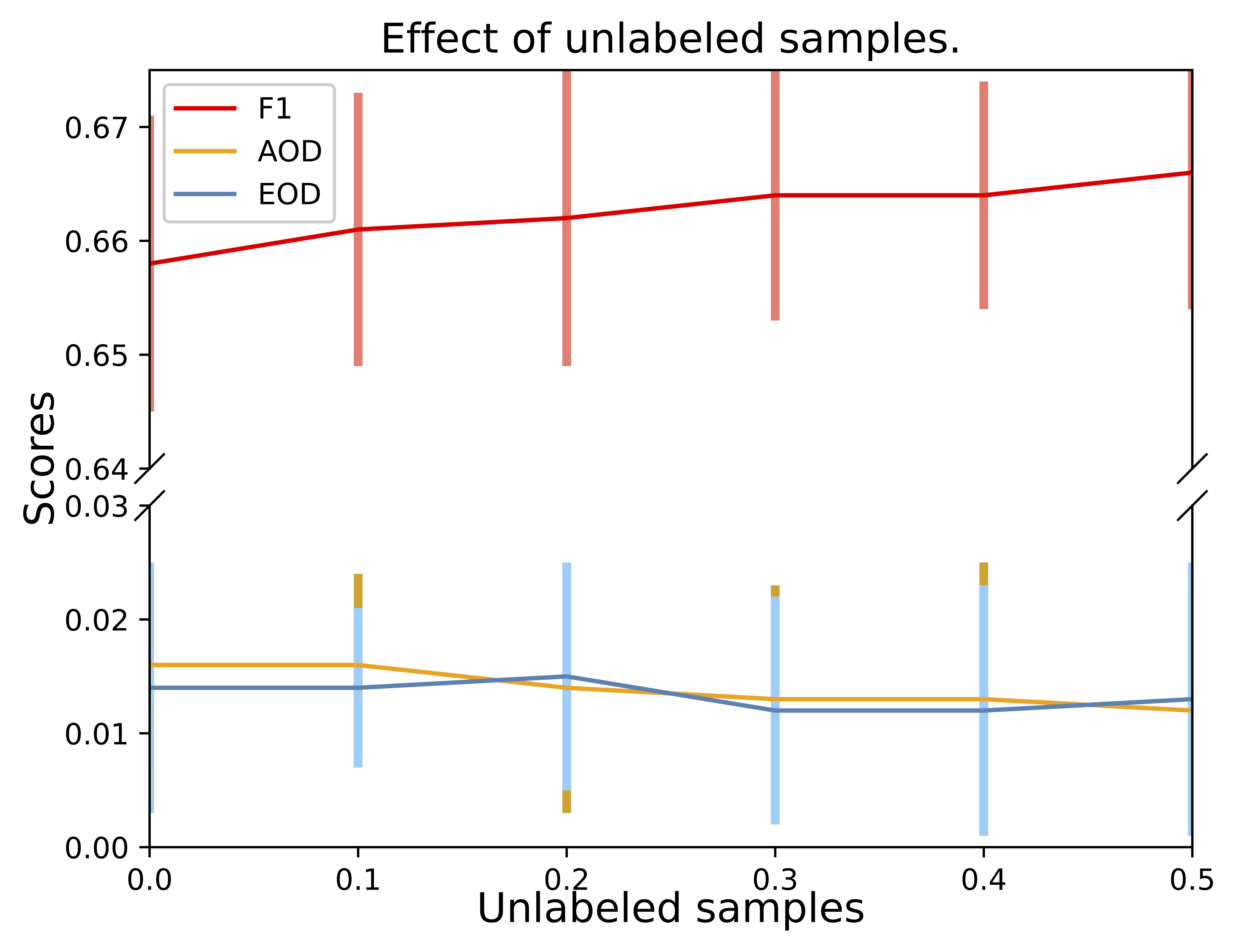}
\caption{F1 scores, AOD and EOD using different numbers of unlabeled examples. Results are averaged over 10 runs on COMPAS.}
\Description{F1 scores, AOD and EOD using different numbers of unlabeled examples.}
\label{fig:unlabel}
\end{figure}
\subsubsection{Effect of unlabeled samples (RQ3)}
 To further study how unlabeled examples affect our framework, we fix the number of positive examples and compare \name's performance with different numbers of unlabeled examples. Since the benchmark data sets are not provided with additional unlabeled data, we deploy the following data generation procedure: we randomly select 50\% examples in the original training set. The remaining 50\% will serve as a pool of unlabeled examples. We test our model leveraging $[0\%, 10\%, 20\%, 30\%, 40\%, 50\%]$ examples in the pool. Note that to ensure the SCAR assumption holds, we randomly label examples of the fixed number from the positive distribution in the current training set.
Average results of 10 runs of \name-EOP on COMPAS are shown in Fig.~\ref{fig:unlabel}. We can see that with more unlabeled examples, our framework achieves better F1 scores and lower AOD and EOD. From the perspective of our framework design, unlabeled examples are leveraged in two aspects: to help predict the labeling probability for building the traditional classifier; to help reduce unfairness in constructing the empirical optimal fair classifier. Adding more unlabeled examples to the training set will benefit both aspects, which is proved by this experiment. This demonstrates \name's advantage in improving both classification performance and fairness simply with unlabeled examples, which may be otherwise useless in other methods.

\section{Conclusion}
\label{sec:con}
In this paper, motivated by real applications such as medical diagnosis and recidivism scoring, we study the important and yet less studied problem of the optimal fair binary classifier in the positive and unlabeled learning using the notion of equalized odds and equal opportunity. In particular, we bridge the gap between fair classification and PUL by first providing the theoretical analysis, and then designing a model-agnostic post-processing framework which preserves favorable consistency properties under mild assumptions. We highlight our framework's flexibility of being easily generalized to any base classifier which outputs conditional labeling probabilities. Extensive experiments demonstrate that our framework outperforms state-of-the-art in both PUL and fair classification. \hide{In future works, we would like to extend our work to other fairness metrics. Moreover, a multi-class extension of PU learning was introduced~\cite{xu2017multi} for applications in predicting multiple classes rather than just positive or negative. The fairness issue within could be explored based on the analysis in our work.}
\begin{acks}
This work is supported by National Science Foundation under Award No. IIS-1947203, IIS-2117902, and IIS-2002540. The views and conclusions are those of the authors and should not be interpreted as representing the official policies of the funding agencies or the government.
\end{acks}
\bibliographystyle{ACM-Reference-Format}
\bibliography{main}

\appendix
\section{Proof of Lemma 1}
\label{prof:lemma1}
\begin{proof}
Using the conditional probabilities, we have
\begin{equation*}
\begin{aligned}
f &= \mathbb{P}(L=1, Y=1 \mid X,S)\\
&=\mathbb{P}(Y=1 \mid X, S)\mathbb{P}(L=1\mid Y=1, X, S)\\
&=\mathbb{P}(Y=1 \mid X, S) \cdot c 
\end{aligned}
\end{equation*}
The result follows by dividing both sides with $c$. 
\end{proof}

\section{Derivation of the Optimal Fair Classifier}
\label{app:deri}
\noindent For EO, using weak duality we can write Equation (\ref{equ:obj}) as:
\begin{equation}
\label{equ:maxmin}
\begin{aligned}
&\min _{g \in \mathcal{G}} \max _{\lambda \in \mathbb{R}}\{\mathcal{R}(g)+\lambda_1(\mathbb{P}(g(X, S)=1 \mid Y=1, S=1)\\
&-\mathbb{P}(g(X, S)=1 \mid Y=1, S=0))\\
 &+\lambda_2(\mathbb{P}(g(X, S)=0 \mid Y=0, S=1)-\mathbb{P}(g(X, S)=0 \mid Y=0, S=0))\}\\
& \geq \max _{\lambda \in \mathbb{R}} \min _{g \in \mathcal{G}}\{\mathcal{R}(g)+\lambda_1(\mathbb{P}(g(X, S)=1 \mid Y=1, S=1)\\
&-\mathbb{P}(g(X, S)=1 \mid Y=1, S=0))\\
 &+\lambda_2(\mathbb{P}(g(X, S)=0 \mid Y=0, S=1)-\mathbb{P}(g(X, S)=0 \mid Y=0, S=0))\}\\
% =:(* *)
\end{aligned}
\end{equation}

In the PUL setting, based on Lemma~\ref{lemma:scar}, we can write:
\begin{equation}
\label{equ:p1p0}
\begin{aligned}
\mathbb{P}(g(X, S)=1 \mid Y=1, S=1)&=\frac{\mathbb{P}(g(X, S)=1, Y=1 \mid S=1)}{\mathbb{P}(Y=1 \mid S=1)}\\
&=\frac{\mathbb{E}_{X \mid S=1}[g(X, 1) f(X, 1)]}{c\mathbb{P}(Y=1 \mid S=1)} \\
\mathbb{P}(g(X, S)=1 \mid Y=1, S=0)&=\frac{\mathbb{P}(g(X, S)=1, Y=1 \mid S=0)}{\mathbb{P}(Y=1 \mid S=0)}\\
&=\frac{\mathbb{E}_{X \mid S=0}[g(X, 0) f(X, 0)]}{c\mathbb{P}(Y=1 \mid S=0)}\\
\mathbb{P}(g(X, S)=0 \mid Y=0, S=1)&=\frac{\mathbb{P}(g(X, S)=0, Y=0 \mid S=1)}{\mathbb{P}(Y=0 \mid S=1)}\\
&=\frac{\mathbb{E}_{X \mid S=1}[(1-g(X, 1)) (1-f(X, 1)/c)]}{\mathbb{P}(Y=0 \mid S=1)} \\
\mathbb{P}(g(X, S)=0 \mid Y=0, S=0)&=\frac{\mathbb{P}(g(X, S)=0, Y=0 \mid S=0)}{\mathbb{P}(Y=0 \mid S=0)}\\
&=\frac{\mathbb{E}_{X \mid S=0}[(1-g(X, 0)) (1-f(X, 0)/c)]}{\mathbb{P}(Y=0 \mid S=0)}
\end{aligned}
\end{equation}

The risk function is:
\begin{equation*}
\begin{aligned}
\mathcal{R}(g) =& \mathbb{P}(g(X, S) \neq Y)\\
=& \mathbb{P}(g(X, S)=0, Y=1)+\mathbb{P}(g(X, S)=1, Y=0) \\
=& \mathbb{P}(g(X, S)=1)+\mathbb{P}(Y=1)-2 \mathbb{P}(g(X, S)=1, Y=1) \\
=& \mathbb{P}(Y=1)+\mathbb{E}[g(X, S)]\\
&-2 \mathbb{E}\left[\mathbf{1}_{\{g(X, S)=1, Y=1\}} \mid S=1\right] \mathbb{P}(S=1) \\
&-2 \mathbb{E}\left[\mathbf{1}_{\{g(X, S)=1, Y=1\}} \mid S=0\right] \mathbb{P}(S=0) \\
=& \mathbb{P}(Y=1)+\mathbb{E}[g(X, S)]\\
&-\frac{2}{c} \mathbb{E}_{X \mid S=1}[g(X, 1) f(X, 1)] \mathbb{P}(S=1) \\
&-\frac{2}{c} \mathbb{E}_{X \mid S=0}[g(X, 0) f(X, 0)] \mathbb{P}(S=0) \\
=& \mathbb{P}(Y=1)-\mathbb{E}_{X \mid S=1}[g(X, 1)(\frac{2}{c} f(X, 1)-1)] \mathbb{P}(S=1) \\
&-\mathbb{E}_{X \mid S=0}[g(X, 0)(\frac{2}{c} f(X, 0)-1)] \mathbb{P}(S=0)
\end{aligned}
\end{equation*}

So the objective function can be simplified as:
\begin{small}
\begin{equation*}
\begin{aligned}
&\mathbb{P}(Y=1)\\
&+\mathbb{E}_{X \mid S=1}[g(X, 1)(f(X, 1)\left(\frac{\lambda_1}{c\mathbb{P}(Y=1 \mid S=1)}-\frac{2}{c} \mathbb{P}(S=1)\right)+\mathbb{P}(S=1)\\
&+ \frac{\lambda_2(f(X,1)/c-1)}{\mathbb{P}(Y=0 \mid S=1)})] \\
&+ \mathbb{E}_{X \mid S=1}\left[\frac{\lambda_2(1-c(X,1)/c)}{\mathbb{P}(Y=0 \mid S=1)}\right] \\
&+\mathbb{E}_{X \mid S=0}[g(X, 0)(f(X, 0)\left(-\frac{\lambda_1}{c\mathbb{P}(Y=1 \mid S=0)}-\frac{2}{c} \mathbb{P}(S=0)\right)+\mathbb{P}(S=0)\\ &+\frac{\lambda_2(1-f(X,0)/c)}{\mathbb{P}(Y=0 \mid S=0)})] \\
&+\mathbb{E}_{X \mid S=0}\left[\frac{-\lambda_2(1-f(X,0)/c)}{\mathbb{P}(Y=0 \mid S=0)}\right] 
\end{aligned}
\end{equation*}
\end{small}

Since $g(X,S) \in \{0,1\}$, we can get the minimizer $g_{\boldsymbol{\lambda}}^*$:
\begin{small}
\begin{equation}
\begin{aligned}
g_{\boldsymbol{\lambda}}^*(X,1) =& \mathbbm{1}_{\{\frac{f(X, 1)}{c}(1-\frac{\lambda_1}{\mathbb{P}(Y=1, S=1)}) + (1-\frac{f(X, 1)}{c}) (1-\frac{\lambda_2}{\mathbb{P}(Y=0, S=1)})\geq 0\}} \\
g_{\boldsymbol{\lambda}}^*(X,0) =& \mathbbm{1}_{\{\frac{f(X, 0)}{c}(1+\frac{\lambda_1}{\mathbb{P}(Y=1, S=0)}) - (1-\frac{f(X, 0)}{c}) (1+\frac{\lambda_2}{\mathbb{P}(Y=0, S=0)})\geq 0\}} 
\end{aligned}
\end{equation}
\end{small}

\section{Proof of Lemma 2}
\label{prof:lemma2}
Substituting the minimizer $g_{\lambda}^*$ into the Lagrange function, we could see that the mappings of $\lambda_1$ and $\lambda_2$ are convex. We can write the first order optimality conditions of the objective function as:

\begin{equation*}
\begin{aligned}
0 \in  &\partial_{\boldsymbol{\lambda}} \mathbb{E}_{X \mid S=1}[g(X, 1)(f(X, 1)\left(\frac{\lambda_1}{c\mathbb{P}(Y=1 \mid S=1)}-\frac{2}{c} \mathbb{P}(S=1)\right)\\
&+\mathbb{P}(S=1)+ \frac{\lambda_2(f(X,1)/c-1)}{\mathbb{P}(Y=0 \mid S=1)})]\\
+&\partial_{\boldsymbol{\lambda}} \mathbb{E}_{X \mid S=1}\left[\frac{\lambda_2(1-c(X,1)/c)}{\mathbb{P}(Y=0 \mid S=1)}\right]\\
+&\partial_{\boldsymbol{\lambda}} \mathbb{E}_{X \mid S=0}[g(X, 0)(f(X, 0)\left(-\frac{\lambda_1}{c\mathbb{P}(Y=1 \mid S=0)}-\frac{2}{c} \mathbb{P}(S=0)\right)\\
&+\mathbb{P}(S=0) +\frac{\lambda_2(1-f(X,0)/c)}{\mathbb{P}(Y=0 \mid S=0)})] \\
+&\partial_{\boldsymbol{\lambda}} \mathbb{E}_{X \mid S=0}\left[\frac{-\lambda_2(1-f(X,0)/c)}{\mathbb{P}(Y=0 \mid S=0)}\right]
\end{aligned}
\end{equation*}

Based on Assumption~\ref{assum:conti}, this subgradient is reduced to the gradient almost surely. So we have Equations (\ref{equ:lambda}):
\begin{equation*}
\begin{aligned}
\frac{\mathbb{E}_{X \mid S=1}[g_{\boldsymbol{\lambda}^*}^*(X,1)f(X, 1)]}{\mathbb{P}(Y=1 \mid S=1)} &= \frac{\mathbb{E}_{X \mid S=0}[g_{\boldsymbol{\lambda}^*}^*(X,0)f(X, 0)]}{\mathbb{P}(Y=1 \mid S=0)} \\
\frac{\mathbb{E}_{X \mid S=1}[(c-f(X, 1))(1-g_{\boldsymbol{\lambda}^*}^*(X,1))]}{\mathbb{P}(Y=0 \mid S=1)} &= \frac{\mathbb{E}_{X \mid S=0}[(c-f(X, 0))(1-g_{\boldsymbol{\lambda}^*}^*(X,0))]}{\mathbb{P}(Y=0 \mid S=0)} 
\end{aligned}
\end{equation*}

Combining it with Equation (\ref{equ:p1p0}), we can see that $\mathbb{P}(g_{\boldsymbol{\lambda}^*}^*(X, S)=1 \mid Y=y, S=1) = \mathbb{P}(g_{\boldsymbol{\lambda}^*}^*(X, S)=1 \mid Y=y, S=0)$. In other words, $g_{\boldsymbol{\lambda}^*}^*$ satisfies  Equalized Odds  with respect to $S$ and is thus fair. Therefore, we have $\mathcal{R}(g_{\boldsymbol{\lambda}^*}^*) \geq \mathcal{R}(g^*)$ because $g^*$ is defined as the optimal fair classifier to minimize the risk. Furthermore, since $(\boldsymbol{\lambda}^*, g_{\boldsymbol{\lambda}^*}^*)$ is a solution to the dual problem, we have  $\mathcal{R}(g_{\boldsymbol{\lambda}^*}^*) \leq \mathcal{R}(g^*)$ according to Equation (\ref{equ:maxmin}). Therefore, we can conclude that $g^* = g_{\boldsymbol{\lambda}^*}^*$.

\section{Proof of Theorem 1}
\label{prof:theo1}
\begin{proof}
Following the strategy of~\cite{denis2020consistency,chzhen2019}, we first introduce an intermediate pseudo-estimator $\Tilde{g}$ as follows:
\begin{equation}
\label{equ:gtilde}
\begin{aligned}
 \Tilde{g}_{\Tilde{\boldsymbol{\lambda}}}(X,1)=& \mathbbm{1}_{\{\hat{p}^+_{X,1}(1-\frac{\Tilde{\lambda_1}}{\mathbb{E}_{X|S=1}[\hat{p}^+_{X,1}]\mathbb{P}(S=1)}) + (1-\hat{p}^+_{X,1}) (1-\frac{\Tilde{\lambda_2}}{(1-\mathbb{E}_{X|S=1}[\hat{p}^+_{X,1}])\mathbb{P}(S=1)})\geq 0\}} \\
 \Tilde{g}_{\Tilde{\boldsymbol{\lambda}}}(X,0)=& \mathbbm{1}_{\{\hat{p}^+_{X,0}(1+\frac{\Tilde{\lambda_1}}{\mathbb{E}_{X|S=0}[\hat{p}^+_{X,0}]\mathbb{P}(S=0)}) - (1-\hat{p}^+_{X,0}) (1+\frac{\Tilde{\lambda_2}}{(1-\mathbb{E}_{X|S=0}[\hat{p}^+_{X,0}])\mathbb{P}(S=0)})\geq 0\}} \\
% \Tilde{g}_{\Tilde{\lambda}}(X,1) =& \mathbbm{1}_{\{1-\hat{p}^+_{X,1}(2-\frac{\Tilde{\lambda}}{\mathbb{E}_{X|S=1}[\hat{p}^+_{X,1}]\mathbb{P}(S=1)})\leq 0\}} \\
% \Tilde{g}_{\Tilde{\lambda}}(X,0) =& \mathbbm{1}_{\{1-\hat{p}^+_{X,0}(2+\frac{\Tilde{\lambda}}{\mathbb{E}_{X|S=0}[\hat{p}^+_{X,0}]\mathbb{P}(S=0)})\leq 0\}} 
\end{aligned}
\end{equation}
where $\Tilde{\boldsymbol{\lambda}}$ satisfies:
\begin{equation}
\begin{aligned}
\label{equ:lambdatilde}
\frac{\mathbb{E}_{X \mid S=1}[\Tilde{g}(X,1)\hat{f}(X, 1)]}{\mathbb{E}_{X|S=1}[\hat{p}^+_{X,1}]} &= \frac{\mathbb{E}_{X \mid S=0}[\Tilde{g}(X,0)\hat{f}(X, 0)]}{\mathbb{E}_{X|S=0}[\hat{p}^+_{X,0}]} \\
\frac{\mathbb{E}_{X \mid S=1}[(\hat{c}-\hat{f}(X, 1))(1-\Tilde{g}(X,1))]}{1-\mathbb{E}_{X|S=1}[\hat{p}^+_{X,1}]} &= \frac{\mathbb{E}_{X \mid S=0}[(\hat{c}-\hat{f}(X, 0))(1-\Tilde{g}(X,0))]}{1-\mathbb{E}_{X|S=0}[\hat{p}^+_{X,0}]} 
\end{aligned}
\end{equation}

Comparing this pseudo-estimator $\Tilde{g}$ with the theoretical ideal $g^*$ in Equation (\ref{equ:optimal}) and our estimator $\hat{g}$ in Equation (\ref{equ:empiricalg}), we can see that $\Tilde{g}$ knows the marginal distribution of $(X,S)$. That is, it has precise information on the distributions $p_S$ and $p_{X|S}$. 
% \he{What do you mean by `knows the marginal distribution'}. 
It can be seen as a nearly-idealized version of $\hat{g}$ where the uncertainty in it is only induced by the estimator $\hat{p}^+_{X,S}$.

To demonstrate our proposed method is asymptotically optimal, we can upper bound the excess risk by expressing it as a sum of two terms, $\mathbb{E}\left[\mathcal{R}(\Tilde{g})\right]-\mathcal{R}(g^*) + \mathbb{E}\left[\mathcal{R}(\hat{g})-\mathcal{R}(\Tilde{g})\right]$. The first term can be bounded by the $l_1$ distance between $\hat{p}^+_{X,S}$ and $p^+_{X,S}$. Based on the first part of Assumption~\ref{assum:cons} that $\hat{p}^+_{X,S}$ is consistent as $n_U, n_L \to \infty$, it converges to zero. For the second term,
% \he{This sentence is confusing.}. 
comparing the upper bound on $\mathcal{R}(\hat{g})$ and the lower bound on $\mathcal{R}(\Tilde{g})$, we can upper bound the difference $\mathcal{R}(\hat{g})-\mathcal{R}(\Tilde{g})$ and show that $\lim_{n_U,n_L \to \infty} \mathbb{E}\left[\mathcal{R}(\hat{g})-\mathcal{R}(\Tilde{g})\right] \to 0$ based on the law of large numbers and the second part of Assumption~\ref{assum:cons}.

To demonstrate our proposed framework is asymptotically fair, we can first upper bound the unfairness with the triangle inequality by considering $TPR^{(1)}$ and $TPR^{(0)}$ respectively and their estimators:
\begin{small}
\begin{equation*}
\begin{aligned}
|TPR^{(1)}-TPR^{(0)}| & \leq |\frac{\mathbb{E}_{X \mid S=1}[\hat{f}(X, 1)\hat{g}(X,1)]}{\mathbb{E}_{X|S=1}[\hat{f}(X,1)]} - \frac{\mathbb{E}_{X \mid S=0}[\hat{f}(X, 0)\hat{g}(X,0)]}{\mathbb{E}_{X|S=0}[\hat{f}(X,0)]}|\\
& + |\frac{\mathbb{E}_{X \mid S=1}[\hat{f}(X, 1)\hat{g}(X,1)]}{\mathbb{E}_{X|S=1}[\hat{f}(X,1)]} - \frac{\hat{\mathbb{E}}_{X \mid S=1}[\hat{f}(X, 1)\hat{g}(X,1)]}{\hat{\mathbb{E}}_{X|S=1}[\hat{f}(X,1)]}|\\
& + |\frac{\mathbb{E}_{X \mid S=0}[\hat{f}(X, 0)\hat{g}(X,0)]}{\mathbb{E}_{X|S=0}[\hat{f}(X,0)]} - \frac{\hat{\mathbb{E}}_{X \mid S=0}[\hat{f}(X, 0)\hat{g}(X,0)]}{\hat{\mathbb{E}}_{X|S=0}[\hat{f}(X,0)]}|
\end{aligned}
\end{equation*}
\end{small}

$|FPR^{(1)}-FPR^{(0)}|$ can be processed in a similar way. We can then prove that $\mathbb{E}\left[\Delta(\hat{g})\right] \leq \mathbb{E}\left[\hat{\Delta}(\hat{g})\right] + o_{U,L}(1)$ using the second part of Assumption~\ref{assum:cons}. Using the consistency and continuity of $\hat{p}^+_{X,S}$ in Assumption~\ref{assum:cons}, and means of theory of empirical processes~\cite{pollard1990empirical}, we can have $\lim \mathbb{E}\left[\hat{\Delta}(\hat{g})\right]$ converges to zero almost surely, which concludes the proof. 
\end{proof}

\section{Implementation Details}
\label{app:hyper}
The three data sets we use can be obtained as follows:
\begin{itemize}
    \item COMPAS is available at https://github.com/propublica/
    
    compas-analysis. 
    \item German Credit is available at https://archive.ics.uci.edu/
    
    ml/machine-learning-databases/statlog/german/.
    \item Drug is available at https://archive.ics.uci.edu/ml/
    
    datasets/Drug+Review+Dataset+\%28Drugs.com\%29
\end{itemize}

The hyper-parameters are found by grid search and we report the best results in terms of classification performance. More details of different models (Lin.LR, Support Vector Machine with polynomial kernel (SVM) and Multilayer Perceptron (MLP)) are shown as follows:
\begin{itemize}
    \item For SVM, $C$ = 0.1, degree = 2, $\gamma$ = 2. 
    \item For Lin.SVM, $C$ = 10, tolerance = 1e-4. 
    \item For Lin.LR, $C$ = 1, solver = 'lbfgs', max iteration=1000. 
    \item For MLP, we use the ReLU activation function and Adam optimizer. For hyper-parameters, regularization term parameter=1e-4, learning rate=1e-3. The hidden layers sizes in MLP are set to $(8, 16, 2)$ for COMPAS, $(24, 48, 2)$ for German Credit and $(12, 24, 2)$ for Drug.
\end{itemize}

For baselines, we use the code provided by original authors and Fairlearn (https://fairlearn.org/), and carefully tune all the hyper-parameters for the best classification performance.

\section{Auxiliary Experimental Results}
\label{app:exp}

{
\begin{table}[h]
\caption{Average results and standard deviation on COMPAS of 10 runs. Labeling rate is 90\%. }
\Description{Average results and standard deviation on COMPAS of 10 runs. }
\centering
\fontsize{7.5pt}{7.5pt}\selectfont
% \vspace{-0.2cm}
\begin{tabular}{c|l|c|c|c} 
\toprule
\multicolumn{2}{c|}{Models}& F1 & AOD & EOD\\
\midrule
\multirow{6}{*}{Lin.LR} & Oracle & 0.621$\pm$0.015 & 0.236$\pm$0.033&0.293$\pm$0.045\\
~ & Na\"ive  &0.552$\pm$0.023 & 0.240$\pm$0.048&0.311$\pm$0.054\\
~& +Agarwal & 0.507$\pm$0.019& 0.032$\pm$0.021 &0.037$\pm$0.036\\
~& +Hardt  &0.357$\pm$0.042 &0.022$\pm$0.012&0.027$\pm$0.019\\
~& +Chzhen &0.491$\pm$0.021 &0.018$\pm$0.015& 0.014$\pm$0.010\\
~ & uPU &0.635$\pm$0.010 &0.075$\pm$0.048&0.096$\pm$0.028\\
~ & wPU & 0.588$\pm$0.036 &0.278$\pm$0.164&0.306$\pm$0.051\\
~ & Bagging & 0.559$\pm$0.020 & 0.293$\pm$0.223&0.335$\pm$0.037\\
~ & \name-EO & 0.656$\pm$0.016&0.015$\pm$0.008&0.013$\pm$0.010\\
~ & \name-EOP &0.658$\pm$0.012 &0.016$\pm$0.007 &0.009$\pm$0.008\\
\midrule
\multirow{6}{*}{SVM} & Oracle  & 0.628$\pm$0.019&0.287$\pm$0.127&0.246$\pm$0.038\\
~ & Na\"ive  & 0.553$\pm$0.022&0.299$\pm$0.102&0.324$\pm$0.073\\
~& +Agarwal  & 0.491$\pm$0.037&0.046$\pm$0.032& 0.035$\pm$0.032\\
~& +Hardt & 0.358$\pm$0.047 &0.028$\pm$0.021& 0.030$\pm$0.023\\
~& +Chzhen  & 0.478$\pm$0.033 &0.032$\pm$0.012&0.040$\pm$0.034\\
~ & uPU  &0.644$\pm$0.012 &0.186$\pm$0.086&0.134$\pm$0.023\\
~ & wPU & 0.565$\pm$0.027 &0.298$\pm$0.076&0.315$\pm$0.055\\
~ & Bagging  &   0.648$\pm$0.013 &0.249$\pm$0.048& 0.286$\pm$0.035\\
~ & \name-EO & 0.646$\pm$0.013&0.012$\pm$0.006&0.015$\pm$0.013\\
~ & \name-EOP &0.648$\pm$0.012&0.016$\pm$0.012&0.012$\pm$0.004\\
\midrule
\multirow{6}{*}{MLP} & Oracle &0.625$\pm$0.008 &0.239$\pm$0.068&0.264$\pm$0.051\\
~ & Na\"ive &0.592$\pm$0.018& 0.248$\pm$0.098&0.299$\pm$0.066\\
~& +Hardt & 0.457$\pm$0.038& 0.036$\pm$0.016&0.029$\pm$0.024\\
~& +Chzhen  &0.548$\pm$0.027 &0.032$\pm$0.029& 0.024$\pm$0.017 \\
~ & uPU  &0.660$\pm$0.012&0.198$\pm$0.087&0.230$\pm$0.067\\
~ & wPU &0.606$\pm$0.021 &0.267$\pm$0.138&0.295$\pm$0.064\\
~ & nnPU & 0.659$\pm$0.023 &0.248$\pm$0.243& 0.289$\pm$0.062\\
~ & \name-EO   &0.663$\pm$0.014 &0.032$\pm$0.017& 0.040$\pm$0.034\\
~ & \name-EOP &0.666$\pm$0.023 &0.036$\pm$0.018&0.034$\pm$0.012\\
% \midrule
% \multicolumn{2}{c|}{nnPU} &  F1 & DEO & F1 & DEO\\
% \midrule
% \multicolumn{2}{c|}{Zafar} &  F1 & DEO & F1 & DEO\\

\bottomrule
\end{tabular}
\label{tab:overall}
\end{table}
}
{
\begin{table}[h]
\caption{Post-processing \name-EOP vs In-processing/Pre-processing on German. The best results for each metric are marked bold.} 
\Description{Post-processing \name-EOP vs In-processing/Pre-processing.} 
\centering
\fontsize{7.5pt}{7.5pt}\selectfont
% \vspace{-0.2cm}
\begin{tabular}{c|c|c|c|c} 
\toprule
Rates& \multicolumn{2}{c|}{Models}  & F1 & EOD\\
\midrule
\multirow{8}{*}{100\%}& \multirow{4}{*}{Lin.SVM} & Na\"ive & 0.565$\pm$0.051 &0.061$\pm$0.042\\
~ & &In &0.541$\pm$0.050 & 0.051$\pm$0.044 \\
~ & &Pre & 0.520$\pm$0.057 & 0.093$\pm$0.069\\
~& &\name & 0.591$\pm$0.035 & 0.056$\pm$0.049\\ 
\cmidrule{2-5}
~& \multirow{4}{*}{Lin.LR} & Na\"ive & 0.557$\pm$0.054& 0.064$\pm$0.053\\
~& &In & 0.540$\pm$0.057& 0.071$\pm$0.064\\ 
~& &Pre & 0.533$\pm$0.057 & 0.062$\pm$0.036 \\ 
~& &\name &0.609$\pm$0.038 & 0.051$\pm$0.041\\ 
\midrule
\multirow{8}{*}{90\%}& \multirow{4}{*}{Lin.SVM} & Na\"ive & 0.471$\pm$0.048 &0.099$\pm$0.052 \\
~ & &In &0.470$\pm$0.033& 0.077$\pm$0.051 \\
~ & &Pre & 0.414$\pm$0.062 & 0.078$\pm$0.074 \\
~& &\name & 0.580$\pm$0.020 &0.034$\pm$0.027 \\
\cmidrule{2-5}
~& \multirow{4}{*}{Lin.LR} & Na\"ive & 0.507$\pm$0.045& 0.132$\pm$0.037\\
~& &In & 0.486$\pm$0.057 & 0.069$\pm$0.044\\
~& &Pre &  0.479$\pm$0.052 & 0.125$\pm$0.072 \\ 
~& &\name &0.606$\pm$0.034 &0.062$\pm$0.047 \\
\bottomrule
\end{tabular}
% \vspace{-0.3cm}
\end{table}
}
{
\begin{table}[H]
\caption{Comparison results of different labeling rates on German.} 
\Description{Comparison results of different labeling rates on German.} 
\centering
\fontsize{7.5pt}{7.5pt}\selectfont
% \vspace{-0.2cm}
\begin{tabular}{c|c|c|c|c} 
\toprule
Rates& \multicolumn{2}{c|}{Models}  & F1 & EOD\\
\midrule
\multirow{4}{*}{100\%}& \multirow{2}{*}{Lin.SVM}& Na\"ive & 0.565$\pm$0.051 &0.061$\pm$0.042\\
~& &\name & 0.591$\pm$0.035 & 0.056$\pm$0.049\\ 
\cmidrule{2-5}
~& \multirow{2}{*}{Lin.LR}& Na\"ive & 0.557$\pm$0.054& 0.064$\pm$0.053\\
~& &\name &0.609$\pm$0.038 & 0.051$\pm$0.041\\ 
\midrule
\multirow{4}{*}{90\%}& \multirow{2}{*}{Lin.SVM}& Na\"ive & 0.557$\pm$0.054& 0.064$\pm$0.053\\
~& &\name & 0.580$\pm$0.020 &0.034$\pm$0.027 \\
\cmidrule{2-5}
~& \multirow{2}{*}{Lin.LR}& Na\"ive & 0.507$\pm$0.045& 0.132$\pm$0.037 \\
~& &\name &0.606$\pm$0.034 &0.062$\pm$0.047 \\
\midrule
\multirow{4}{*}{80\%}& \multirow{2}{*}{Lin.SVM}& Na\"ive &0.193$\pm$0.136 &0.044$\pm$0.051\\
~& &\name &0.527$\pm$0.047 &0.033$\pm$0.029\\
\cmidrule{2-5}
~& \multirow{2}{*}{Lin.LR}& Na\"ive &0.401$\pm$0.046 &0.083$\pm$0.055\\
~& &\name &0.605$\pm$0.027&0.056$\pm$0.040\\
\midrule
\multirow{4}{*}{50\%}& \multirow{2}{*}{Lin.SVM}& Na\"ive & 0.000$\pm$0.000&0.000$\pm$0.000\\
~& &\name & 0.481$\pm$0.042 &0.019$\pm$0.018 \\
\cmidrule{2-5}
~& \multirow{2}{*}{Lin.LR}& Na\"ive &0.106$\pm$0.036&0.053$\pm$0.026\\
~& &\name &0.593$\pm$0.036&0.044$\pm$0.019\\
\bottomrule
\end{tabular}
\end{table}
}

\end{document}